\documentclass[preprint,12pt]{elsarticle}

\usepackage{amssymb}
\usepackage{amsmath}

\usepackage{xcolor}
\newcommand{\sjw}[1]{{\textbf{\textcolor{red}{#1}}}}

\usepackage{amsthm}
\usepackage{amsmath,amssymb,enumerate}
\usepackage{mathrsfs}
\usepackage{times}
\usepackage{multicol}
\usepackage{amsfonts}
\usepackage{textcomp}
\usepackage{balance}
\usepackage{multirow}
\usepackage{booktabs}
\usepackage{dblfloatfix} 
\usepackage{bbold}
\usepackage{makecell}
\usepackage{hhline} 
\usepackage{float} 
\usepackage[mathscr]{euscript}

\usepackage{comment}
\usepackage{xparse}
\usepackage{lipsum}
\usepackage{soul} 
\usepackage{balance}
\usepackage{dsfont} 
\usepackage{setspace} 
\usepackage{accents} 

\usepackage{graphicx}
\usepackage{fixltx2e}

\usepackage[T1]{fontenc}
\usepackage[usenames,dvipsnames,svgnames,table, xcdraw]{xcolor}
\usepackage[caption=false,font=footnotesize]{subfig}
\usepackage[utf8]{inputenc}

\usepackage{hyperref}
\hypersetup{
  colorlinks=true,pdfpagemode=UseNone,citecolor=black,linkcolor=black,urlcolor=BrickRed
}

\usepackage{caption}
\graphicspath{
    {fig/}
}

\usepackage[ruled]{algorithm2e}
\newcommand{\nextnr}{\stepcounter{AlgoLine}\ShowLn}
\usepackage{algpseudocode}
\algnewcommand\algorithmicforeach{\textbf{for each}}
\algdef{S}[FOR]{ForEach}[1]{\algorithmicforeach\ #1\ \algorithmicdo}
\SetKwRepeat{Do}{do}{while}
\SetKwInput{KwInput}{Input}                
\SetKwInput{KwOutput}{Output}              

\makeatletter
\newcommand{\mathleft}{\@fleqntrue\@mathmargin0pt}
\newcommand{\mathcenter}{\@fleqnfalse}
\makeatother

\journal{Submitted to Robotics and Autonomous Systems}

\begin{document}

\begin{frontmatter}



\title{DynaWeightPnP: Toward global real-time 3D-2D solver in PnP without correspondences} 


\author{Jingwei Song$^1$ and Maani Ghaffari$^2$} 

\affiliation{organization={1.United Imaging Research Institute of Intelligent Imaging, Beijing 100144, China},
            city={Beijing},
            postcode={100144}, 
            country={China}}
\affiliation{organization={2.University of Michigan},
            city={Ann Arbor},
            postcode={48109}, 
            state={MI},
            country={USA}}

\begin{abstract}
This paper addresses a special Perspective-n-Point (PnP) problem: estimating the optimal pose to align 3D and 2D shapes in real-time without correspondences, termed as correspondence-free PnP. While several studies have focused on 3D and 2D shape registration, achieving both real-time and accurate performance remains challenging. This study specifically targets the 3D-2D geometric shape registration tasks by leveraging the recently developed Reproducing Kernel Hilbert Space (RKHS) registration framework to address the partial overlap challenge. An iteratively reweighted least squares method is designed to solve the RKHS-based formulation efficiently. Moreover, our work identifies a unique and interesting observability issue in correspondence-free PnP: the numerical ambiguity between rotation and translation. To address this, we proposed DynaWeightPnP, introducing a dynamic weighting sub-problem and an alternative searching algorithm to enhance pose estimation and alignment accuracy. Experiments were conducted on a 3D-2D vascular centerline registration task within Endovascular Image-Guided Interventions (EIGIs). Results demonstrated that the proposed algorithm achieves registration processing rates of 60 Hz (without post-refinement) and 31 Hz (with post-refinement) on modern single-core CPUs, with competitive accuracy comparable to existing methods. 
\end{abstract}


\begin{highlights}
\item It reveals the main reason for the numerous local minima observed and noted as a limitation in previous work. We qualitatively identify that this phenomenon is due to the ambiguity between rotation and translation in correspondence-free PnP, an issue unique to this scenario. 
\item We introduce a novel structure, DynaWeightPnP, which incorporates a reweighted subproblem and alternative searching into previous work. DynaWeightPnP only requires an additional $100\%$ of the computational time while reducing the registration error by $20\%$ to $70\%$. 
\item We conduct more synthetic and in-vivo experiments to support the claims made in this article. 
\end{highlights}


\end{frontmatter}




\section{Introduction}

    Shape registration is the process of aligning multiple geometric shapes of the same object observed from different viewpoints, potentially varying in time, field of view, sensor, and dimension. Providing correct viewing poses and shape alignments serves as a fundamental technique across various fields such as computer vision (3D reconstruction and panorama stitching), robotics (localization), cartography (surveying and mapping), and medicine (multi-modal medical image processing)~\cite{zitova2003image}. Classified by correspondence and shape dimension, shape registration problems typically fall into the following categories: correspondence-based 3D-3D, correspondence-free 3D-3D, correspondence-based 3D-2D, and correspondence-free 3D-2D problems. Correspondence-based 3D-3D methods offer robust initial registrations, while correspondence-free 3D-3D techniques, such as those in the Iterative Closest Point (ICP) family, support sequential Visual Odometry (VO) tasks. Correspondence-based 3D-2D methods, belonging to the Perspective-n-Point (PnP) family, localize the pose associated with the 2D shape. The three categories have been extensively researched and applied in industry. Surprisingly, correspondence-free 3D-2D methods have not received extensive analysis to date. 

    \begin{figure}[t]
        \centering
        \includegraphics[width=1\columnwidth]{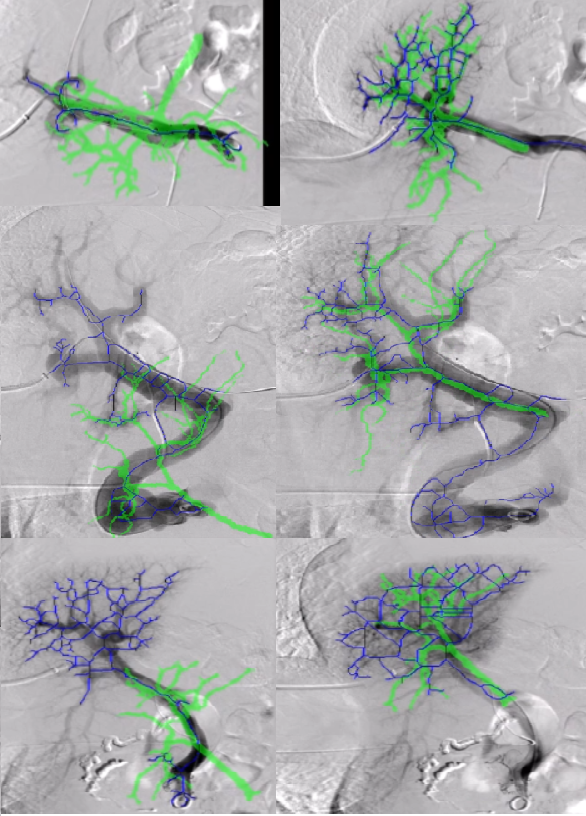}
        \caption{The figure illustrates sample results of the proposed DynaWeightPnP method for 3D-to-2D vessel registration. The 3D vessel is represented by a segmented mesh from CTA, while the 2D vessel data is derived from DSA. The left and right columns display the results before and after applying the correspondence-free PnP approach, respectively.}
        \label{Fig_3D_2D_sample}
    \end{figure}

    
    The limited research focusing on correspondence-free 3D-2D alignment approaches may be attributed to the fact that correspondences between 3D and 2D data can be easily obtained, especially due to well-preserved texture information in most 2D images. For instance, in image-based localization tasks~\cite{tian20203d,song2022fusing}, shape registration and pose estimation rely heavily on descriptor matching of 3D-2D corner points. Off-the-shelf descriptors like SIFT, SURF, BRISK, FAST, and ORB, which offer translation, rotation, and illumination invariant features, facilitate high-quality key point registration, enabling PnP methods to achieve precise alignment. To our knowledge, correspondence-free 3D-2D matching algorithms are narrowly applied in tasks such as edge-based textureless object 6 DoF (silhouette shape) registration~\cite{harris1990rapid,drummond2002real,choi20123d,seo2013optimal,wang2020multientity,dong2020accurate,tian2022large,perez2024alignment} or multi-modal image registration where aligning structures based on texture information alone is challenging~\cite{groher2009deformable,labrunie2022automatic}, thus limiting their applications. However, with the rapid advancements in special robots like intelligent medical and surgical robots, real-time multi-modal data registration has gained attention as it forms the foundation for instrument localization and robot manipulation. Take Endovascular Image-Guided Interventions (EIGIs) as an example, where intra-operative fluoroscopic X-ray imaging or Digitally Subtracted Angiograms (DSA) often need to be aligned with pre-operative Computed Tomography Angiography (CTA) or 3D DSA. Unlike typical computer vision tasks where 3D and 2D data exhibit high consistency in color and illumination spaces, EIGIs face significant challenges introduced by different imaging systems capturing the same target across imaging domains. Therefore, extracting semantic 3D and 2D centerlines as invariant features and feeding them into correspondence-free PnP methods becomes crucial. Fig. \ref{Fig_3D_2D_sample} shows an example of real-time CTA to DSA image registration. This research aims to systematically review existing correspondence-free PnP methods, predominantly proposed for EIGIs tasks and introduces the prior-free DynaWeightPnP as a robust and accurate approach to solving correspondence-free PnP problems.\par 

    Similar to other approaches for shape alignment, correspondence-free 3D-2D alignment methods can also be categorized into model-driven and data-driven approaches. \cite{mitrovic20183d} provided a comprehensive review and categorized model-driven approaches based on their similarity measurement metrics: intensity-based, feature-based, and gradient-based. Intensity metrics refer to normalized illumination metrics like the Structural Similarity Index Measure. Feature metrics involve invariant semantic features such as vessel centerlines. Gradient-based metrics aim to align 3D shapes to 2D data by minimizing the sum of gradients. In contrast to manually crafted similarity metrics in conventional approaches, data-driven methods employ Deep Neural Networks (DNNs), which can encode prior information and learn metrics implicitly. Existing data-driven methods~\cite{zheng2018pairwise, liao2017artificial, miao2018dilated} model the entire alignment procedure by predicting rigid spatial transformations from 3D and 2D pairs in an end-to-end manner. For instance, \cite{zheng2017learning} proposed an end-to-end shape-to-pose estimation approach. Subsequently, \cite{zheng2018pairwise, miao2018dilated, miao2019agent} extended this framework with reinforcement learning for iterative alignment. Since the objective of this work is to propose a general and prior-free approach, DNN-based methods are beyond its scope.\par 
    
	This work summarizes that existing correspondence-free 3D-2D registration algorithms mainly follow an iterative PnP procedure, which involves iteratively aligning points by their closest point in 2D space and then conducting PnP. However, there are \textbf{two major obstacles} to naively implementing correspondence-free PnP: the \textbf{partial overlap} problem and the \textbf{numerical observability} issue between rotation and translation. Partial overlap refers to the wrong alignment from a 3D model to a 2D model that cannot be fully aligned. This misalignment is caused by missing or redundant features from the \textbf{partial overlapping}. Since the 2D shape lacks the correct 3D scale, the 3D shape can adjust the 3D viewing depth (Z-axis in the pin-hole camera model) and enforce the scale of the "shrunk" projected 3D model to match the 2D model. Another interesting issue is \textbf{numerical observability}, which refers to the fact that rotation and translation are not separable when solving the correspondence-free PnP problem. As the target 2D data lacks the Z dimension, the numerical observability issue introduces a significant number of local minima, degrading existing algorithms' performance. Additionally, in EIGIs specifically, other issues include heavy 3D and 2D shape noise due to segmentation uncertainty, excessive outliers, and non-rigid deformation. This article handles all the above-mentioned issues except non-rigid deformation modeling. Interested readers are encouraged to refer to previous works~\cite{rivest2012nonrigid,song2023iterative} for non-rigid formulation.\par

    To handle the partial overlap problem and numerical observability issue, this work extends our previous work~\cite{song2023iterative}, which used a 2D Reproducing Kernel Hilbert Space (RKHS) formulation, an Iterative Reweighted Least Squares (IRLS) solver, and an analytical second-order derivative-based solver for real-time and partial overlap shape matching. DynaWeightPnP, as an extension of~\cite{song2023iterative} incorporates a reweighted subproblem and an alternative searching strategy. DynaWeightPnP significantly reduces the local minima presented in the previous method~\cite{song2023iterative}. In summary, the contributions of this research are as follows. 

    \begin{enumerate}
        \item Our proposed DynaWeightPnP employs an RKHS loss and in conjunction with IRLS solver. Comparing to the conventional methods with Euclidean distance and Huber loss, it handles the partial overlap issue, and significant enhance the registration robustness. 
        \item We discover an interesting and unique issue in correspondence-free PnP, which is the numerical rotation and translation ambiguity. The numerical observability issue introduces a large number of local minima, hindering the performance of existing algorithms.
        \item To tackle the numerical observability issue, our proposed DynaWeightPnP incorporates a reweighted subproblem and an alternative searching strategy. DynaWeightPnP mitigates the impact of local minima, yielding more accurate pose and registration.
    \end{enumerate}

    This article extends our preliminary work~\cite{song2023iterative} and differs in the following ways. First, it reveals the main reason for the numerous local minima observed and noted as a limitation in~\cite{song2023iterative}. We qualitatively identify that this phenomenon is due to the ambiguity between rotation and translation in correspondence-free PnP, an issue unique to this scenario. Second, we introduce a novel structure, DynaWeightPnP, which incorporates a reweighted subproblem and alternative searching into~\cite{song2023iterative}. DynaWeightPnP only requires an additional $100\%$ of the computational time while reducing the registration error by $20\%$ to $70\%$. Third, we conduct more synthetic and in-vivo experiments to support the claims made in this article. Lastly, the non-rigid formulation is not discussed in this research.\par


	\section{Related works}
	\label{section_relatedworks}

    
    While correspondence-free PnP appears straightforward given the presence of PnP and ICP, it is surprisingly under-discussed. This can be attributed to the ease of obtaining robust correspondences between 3D and 2D textured natural images. In contrast to the process of matching natural images, where correspondences are readily available, multi-modal image matching algorithms unify images in the same domain, utilizing segmented geometric shapes and establishing matches based on geometric similarity, as pointed out in \cite{jiang2021review}. 3D-2D correspondence-free PnP algorithms resemble existing 3D-3D point cloud matching methods and can be categorized as model-driven and data-driven approaches.\par

    Model-driven approaches formulate the problem as authentic PnP~\cite{aylward2003registration,markelj2008robust,rivest2012nonrigid} with three main differences. First, Closest Point Searching~\cite{besl1992method,drummond2002real} (CPS) or contour directional-based searching~\cite{huang2021pixel,tian2022large} was applied to build temporal data associations. Second, the objective function was solved iteratively, allowing data re-association and refinement, unlike P3P~\cite{gao2003complete,Lepetit_160138} or EPnP that aimed for closed-form solution. Existing model-driven approaches vary in the loss function and the optimizer. \cite{drummond2002real,groher2009deformable,rivest2012nonrigid} retrieved the optimal 6 Degree-of-Freedom (DoF) transformation by minimizing the 2D Euclidean sum loss. Although deformation parameters were also estimated in these works, they are beyond the scope of this discussion. Different from Euclidean loss, \cite{aylward2003registration} obtained the 6 DoF transformation by maximizing the sum of Gaussian kernel in 2D space and measuring the similarity between the 2D and projected 3D shape. \cite{mitrovic20133d} used a derivative-free algorithm (Powell) for Euclidean loss sum loss and implemented it on the GPU-end for fast optimization. \cite{groher2009deformable} adopted the derivative-based algorithm Broyden–Fletcher–Goldfarb–Shanno (BFGS) as the solver. BFGS used pseudo 2D Hessian instead of expensive Hessian calculation. \cite{aylward2003registration} maximized the sum Gaussian kernel with gradient ascend, and its first-order solver is slow. Three algorithms~\cite{jomier20063d,rivest2012nonrigid,mitrovic20133d} achieved around $0.5$s while the rest require over $10$s. Although not reported explicitly, \cite{jomier20063d} is expected to be slow because it is based on \textit{genetic algorithm}. ~\cite{song2023iterative} adopted 2D RKHS space and imported IRLS solver for the analytical second-order derivative-based solver. Among existing model-driven approaches, \cite{song2023iterative} achieved the highest accuracy and fastest speed. Lastly, researches for 6 DoF object tracking reported local minima issues. To handle the local minima, particle filters algorithm~\cite{choi20123d,seo2013optimal,tian2022large} is widely used to jump out of local minima.\par

    Data-driven algorithms leverage DNN to learn the similarity metric and determine the optimal 6 DoF transformation. Unlike handcrafted metrics such as Gaussian kernels or Euclidean loss, coupled with analytical optimizers, data-driven algorithms encode prior knowledge and execute processes in an end-to-end manner. In practice, a vast number of 2D projections are simulated with the 3D shape and selected transformations (label) in order to produce labeled training data for model pre-training. The pre-trained model is optionally and fine-tuned on a real-world labeled data set. Early research~\cite {zheng2018pairwise,zheng2017learning} used multiple networks for sensor pose estimation in a coarse-to-fine manner. Each network covers a certain range (scale) of the pose. \cite{guan2019deformable} proposed a multi-channel convolution DNN that integrates multiple phases caused by the breathing and heartbeat of patients. Its inference step consumes 4 milliseconds, thanks to the small size of the neural network. With the development of RL, the following research attempts to align 3D-2D shapes based on intelligent RL agents. The trained agent searches for the optimal transformation iteratively. \cite{miao2018dilated,miao2019agent} adopted a simplified Q-learning based on the Markov decision process. \cite{guan2020transfer} adopted transfer learning to bridge the domain gap between training and testing data sets. In contrast to prior-free model-based methods, learning-based approaches' performance requires a massive amount of training data and high-quality sim-to-real model transferring.

    This research aims at general real-time 3D-2D registration algorithms primarily intended for robot navigation applications. Consequently, we only focus on model-driven methods, given their higher level of explicability and analyzability. In contrast, data-driven approaches are better suited for specific predefined tasks where there is consistency in data distribution between training and testing data sets. As~\cite{meng2022weakly} pointed out, these complicated DNN-based approaches suffer from ``hardness of obtaining the ground-truth transformation parameters'' and ``real intra-operative DSAs can be incomplete''.\par

    \begin{figure}[t]
        \centering
        \subfloat{
            \begin{minipage}[]{0.9\textwidth}
                \centering
                \includegraphics[width=1\linewidth]{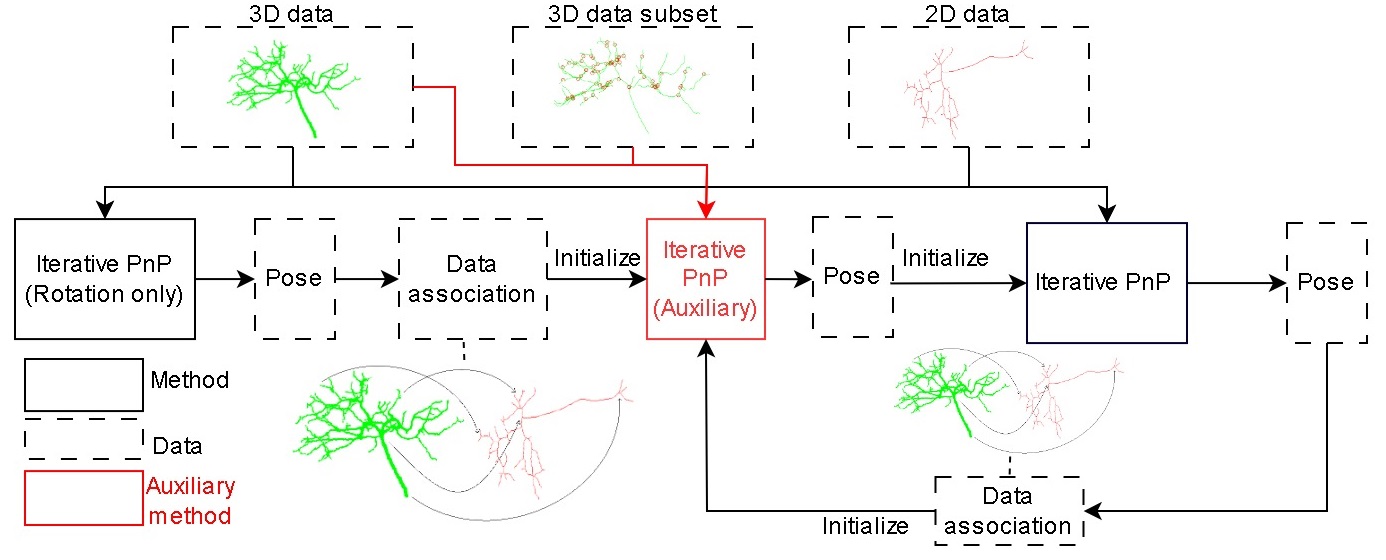}
            \end{minipage}
        }
        \caption{Illustrated is the procedure of our proposed DynaWeightPnP. It corresponds to the iterative searching structure shown in Algorithm \ref{Algorithm_ADMM}.}
        \label{fig_framework}
    \end{figure}

    \section{Methodology}
    \label{secion_methology}

    This section establishes the theoretical and algorithmic foundation of our correspondence‑free PnP framework. Our proposed framework can be summarized as Fig.~\ref{fig_framework}. We begin by reviewing the classical PnP formulation and extending it to the correspondence‑free setting, identifying core challenges such as partial overlap and local minima (Section~\ref{subsecion_PnP_corresFreePnP}). Next, we introduce a RKHS representation combined with IRLS to achieve robust, real‑time optimization under noisy and incomplete observations (Section~\ref{Sec_partial_overlap_rkhs}). Finally, we analyze fundamental numerical observability issues—including scale ambiguity and rotation–translation coupling—and propose a dynamic re‑weighting strategy with alternating optimization to escape poor local minima and improve pose accuracy (Section~\ref{Sec_dyna_weight_search}).

	\subsection{PnP and correspondence-free PnP}
    \label{subsecion_PnP_corresFreePnP}
    
    \textbf{The general principle of PnP}. PnP is the process of finding the optimal 6 DoF spatial transformation that aligns a set of 3D points to 2D points on condition that the 3D-to-2D projection matrix has been well calibrated and 3D-to-2D correspondences are known~\cite{fischler1981random}.
    Define the optimal camera (sensor) pose as $\mathbf{T} \in \mathrm{SE}(3)$. $\mathbf{T}$ aligns the moving 3D point $\mathbf{p}_i \in \mathbb{R}^4$ ($i \in \Omega$ and $\Omega$ is the point index set) and the fixed 2D point $\mathbf{q}_i \in \mathbb{R}^3$. Denote $\{\mathbf{p}\}$ and $\{\mathbf{q}\}$ as the sets of $\mathbf{p}_i$ and $\mathbf{q}_i$. Please note that this article represents all 3D and 2D points in homogeneous coordinate form. Authentic PnP is formulated as solving the optimal pose $\mathbf{T}^*$ by minimizing\par
\begin{equation}
	\label{Eq_pnp}	\mathbf{T}^*=\operatorname{argmin}_{\mathbf{T}} \sum_{i \in \Omega}||\pi(\mathbf{T}\mathbf{p}_i,\mathbf{K})-\mathbf{q}_i||^2_2,
\end{equation}

 \noindent where $\pi(\mathbf{T}\mathbf{p}_i,\mathbf{K})$ denotes the projection matrix that maps the 3D shape to 2D plane based on the calibrated camera intrinsic matrix $\mathbf{K} \in \mathbb{R}^{3\times4}$ and the pin-hole camera projection function defined as
\begin{equation}
\pi(\mathbf{T}\mathbf{p}_i,\mathbf{K}) = \mathbf{K}\mathbf{T}\mathbf{p}_i.
\end{equation}

\textbf{Correspondence-free PnP: the PnP without predefined correspondences}. As the name suggested, correspondence-free PnP is a general algorithmic paradigm that alternates between determining the correspondences and solving an authentic PnP problem~\cite{haralick1994review,Lepetit_160138}. Unlike traditional PnP, correspondence-free PnP does not require predefined correspondences. To address this issue, it employs repetitive CPS and PnP to obtain the optimal 3D transformation $\mathbf{T}^*$. In each step $k$, denote $\mathbf{q}^{k}_i \in {\mathbf{q}}$ as the closest point to $\mathbf{p}_i$ (given the current transformation $\mathbf{T}^{k}$). The optimal estimated pose for the next step is denoted as $\mathbf{T}^{k+1}$, with $k$ starting from 1. Similar to \eqref{Eq_pnp}, \cite{rivest2012nonrigid}~defines 2D points' Euclidean distance sum as the loss function and obtain optimal sensor pose as
\begin{equation}
	\label{2Ddis_object_func_0_0}	\mathbf{T}^{k+1}=\operatorname{argmin}_{\mathbf{T}} \sum_{\mathbf{p}_i \in \{\mathbf{p}\}, \mathbf{q}^{k}_i \in \{\mathbf{q}\}}||\pi(\mathbf{T}\mathbf{p}_i,\mathbf{K})-\mathbf{q}^{k}_i||^2_2,
\end{equation}

\noindent where $\mathbf{T}$ is initialized with $\mathbf{T}^{k}$ in step $k$. Instead of known correspondences, the temporal correspondences between the source set $\{\mathbf{p}\}$ and target set $\{\mathbf{q}\}$ are determined by 2D CPS. $\mathbf{q}^{k}_i$ is the closest point to $\mathbf{p}_i$ on 2D plane given $\mathbf{T}^k$. \cite{rivest2012nonrigid} relaxed the rotation in $\mathbf{T}$ as an affine matrix for modelling simple deformation. Furthermore, unlike authentic PnP in \eqref{Eq_pnp}, \eqref{2Ddis_object_func_0_0} is very dependent on the initial input $\mathbf{T}^{0}$.\par

\begin{figure}[t]
    \centering
    \includegraphics[width=0.95\columnwidth]{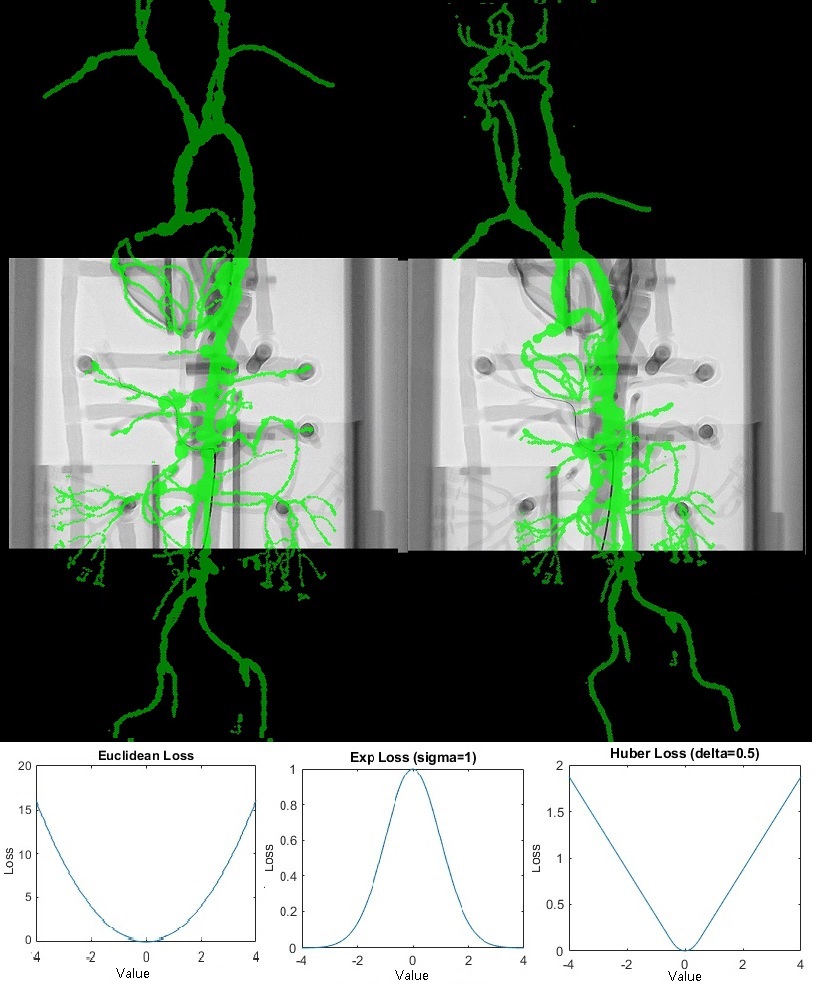}
    \caption{The upper two figures show the typical \textbf{partial overlap} shape alignment. The green shape is the projection of the pre-operative 3D vessel, which is in large size. The upper left figure is the proposed correspondence-free PnP on RKHS space, which is robust to \textbf{partial overlap} shape alignment. The upper right figure is our implementation of ~\cite{rivest2012nonrigid}, which uses the Huber loss. The bottom figures are three loss functions. Only the ``Exp'' (stands for exponential) loss manages to lower the impacts of ``outlier'' points.}
    \label{fig_big_to_small}
\end{figure}

Equation~\eqref{2Ddis_object_func_0_0} with iterative CPS can be summarized as \textbf{correspondence-free PnP}. Correspondence-free PnP has seldom been discussed in robotic and computer vision communities because 3D-2D correspondences can be obtained with well-known image feature point matching algorithms, making traditional PnP more suitable. However, to our knowledge, correspondence-free PnP is popular in multi-modal data registration applications, such as in the medical domain. For example, \cite{rivest2012nonrigid} demonstrates that the CTA-generated 3D coronary artery and fluoroscopy-generated 3D coronary artery cannot be associated in the texture domain. Instead, correspondence-free PnP acquires the optimal registration in the geometry, which bridges both domains.\par 

Our previous research revealed that correspondence-free PnP suffers two issues: \textbf{partial overlap} and \textbf{local minima}~\cite{song2023iterative}. \textbf{partial overlap} is a notorious issue in correspondence-free point cloud registration algorithms~\cite{arnold2021fast}, mostly due to the limited field of view of the sensors. Thus, correspondence-free PnP~\cite{rivest2012nonrigid} inevitably suffers from incorrect registration in contrast to authentic PnP. Fig. \ref{fig_big_to_small} shows a typical \textbf{partial overlap} shapes alignment (``big-to-small''), where the green vessel is the projected 3D object which is much larger than the target 2D object. The incorrect registration, a special form of an outlier, is the major source of inaccuracy in PnP~\cite{arnold2021fast}. Multi-stage methods~\cite{gao2003complete,kneip2011novel,hu2002note,quan1999linear,lepetit2009ep,li2012robust} use a lightweight PnP solver (P3P, P4P and P5P in most cases) combined with RANSAC~\cite{fischler1981random} to remove outliers. \par 

Meanwhile, our previous work~\cite{song2023iterative} experimentally revealed another interesting phenomenon that ``point-wise registration is still successful even though the estimated pose is wrong''. It means correspondence-free PnP suffers far more potential local minima than PnP and even numerical observability issues.

\subsection{Partial overlap and RKHS}
\label{Sec_partial_overlap_rkhs}
This section adopts the idea of 2D RKHS~\cite{clark2021nonparametric} for robustness and IRLS~\cite{ray2024rkhsba} for fast estimation. 

\subsubsection{RKHS space}

RKHS is a Hilbert space (complete metric space) of functions in which point evaluation is a continuous linear functional~\cite{berlinet2011reproducing}. The representer theorem states that the linear combination of the kernel functions $\operatorname{K}(\mathbf{x}_i,\cdot)$ ($\mathbf{x}_i$ the sample in low dimension) fits the high dimensional function and kernel trick ensures the point-wise estimation in low dimension~\cite{mohri2018foundations}. Therefore, RKHS is widely used in machine learning algorithms like Support Vector Machines for fitting high dimensional complicated functions~\cite{steinwart2008support}. \par 

Continuous Visual Odometry (CVO)~\cite{clark2021nonparametric} is the first work applying RKHS theory in point cloud registration. RKHS converts the point clouds into a high-dimensional smooth feature space in comparison to the widely used 3D Euclidean distance in the ICP problem. Gaussian kernel coincides with Gaussian kernels or Gaussian Mixture Model (GMM)~\cite{biber2003normal,magnusson2007scan,jian2010robust}, is employed as the kernel function. GMM-based approaches are in discrete Gaussian kernel space, while CVO is continuous and equipped with an inner product metric.\par 

This research follows CVO~\cite{clark2021nonparametric,ray2024rkhsba}'s 3D point cloud registration pipeline by adopting RKHS in handling \textbf{partial overlap} (or ``big-to-small'') issue in correspondence-free PnP scenarios. The Euclidean loss in the authentic formulation in \eqref{2Ddis_object_func_0_0} brings discontinuity and outliers. As Fig. \ref{fig_big_to_small} illustrates, many unpaired source points are wrongly matched to target points, leading to erroneous registration, especially in terms of scale. It should be noted that cross-searching for one-to-one correspondences is not applicable for real-world robots because the fast spatial searching approach KD tree~\cite{muja2009flann} can only be built on the static 2D points $\{\mathbf{q}\}$. Without a KD tree, real-time searching in 2D space can hardly be achieved. M-estimators~\cite{maronna1976robust} and thresholding can mitigate this issue, but both are sensitive to parameter settings and scenarios. RKHS, particularly in the Gaussian kernel, can elegantly handle both discontinuity and outliers in correspondence-free PnP. Regarding discontinuity, the RKHS field is a sum of kernel values and is continuous everywhere in the RKHS space. As to outliers, the Gaussian kernel (shown in the bottom of Fig. \ref{fig_big_to_small}) assigns a very small weight to registration at a very large distance. In such manner, the outliers (the example in \textbf{``big-to-small''}) have a very limited impact on final registration. Following \cite{clark2021nonparametric}, correspondence-free PnP in RKHS loss can be formulated as

\begin{equation}
\begin{aligned}
    \label{GMM_object_func}
    &
 \mathbf{T}^{k+1}=\operatorname{argmax}_ {\mathbf{T}}  E_{data} , \text{ such that},
 \operatorname{min}_ {\mathbf{T}} E_{init}\\
 &E_{data}= \sum_{\mathbf{p}_i \in \{\mathbf{p}\}}\mathrm{w}_i\sum_{\mathbf{q}^{k}_i \in \{\mathbf{q}\}}\exp \left(-\frac{\lVert \pi(\mathbf{T}\mathbf{p}_i,\mathbf{K})-\mathbf{q}^k_j \rVert^2_2}{2\ell^2}\right)\\
  &E_{init}= \lVert \operatorname{log}(\mathbf{T}^{-1}_0\mathbf{T})^{\vee} \rVert_2^{2},
\end{aligned}
\end{equation}

\noindent where $\mathrm{w}_i$ is the weight and uniformly set as 1 in this research, $\mathbf{T}_0$ is the initial pose collected from sensor calibration, and $\ell$ is the scale parameter. $(\cdot)^\vee$ convert Lie algebra $\mathfrak{se}(3)$ to 6-vector $\mathbb{R}^6$ and defines the distance on $\mathrm{SE}(3)$ manifold. \eqref{GMM_object_func} is solved iteratively by searching corresponding $\mathbf{q}^{k}_i$ for $\mathbf{p}_i$ in each iteration as CPS. It should be pointed out that $E_{init}$ is adopted because $\mathbf{T}^*$'s z-direction may increase enormously, and the 3D shapes' projection shrinks to a point and matches an arbitrary target point. This phenomenon represents one scenario of outlier and thus should be handled with $E_{init}$. The scale parameter $\ell$ is defined as the maximum $||\pi(\mathbf{T}\mathbf{p}_i,\mathbf{K})-\mathbf{q}_j||^2_2$ and shrinked by half every 5 iterations as coarse-to-fine registration. Please note that \eqref{GMM_object_func} suffers from constraint $\operatorname{min}_ {\mathbf{T}, \theta} E_{init}$ and only has Pareto optimum (infinite solutions, as improving one objective necessarily worsens the other without a predefined preference). Next section adopted IRLS~\cite{ray2024rkhsba} for best gradient searching and energy-constraint balancing.\par 

By comparing with authentic CVO works~\cite{clark2021nonparametric,ray2024rkhsba}, readers may notice that \eqref{GMM_object_func} employs $E_{init}$ to constrain that the estimated pose does not deviate much from the initial pose $\mathbf{T}_0$. \sjw{This is because correspondence-free PnP is numerically unobservable due to the lack of the Z-dimension. And both the $\mathrm{SE}(3)$ manifold and sum of Gaussian kernels introduce nonconvexity to the objective function.} Therefore, $E_{init}$ is used to provide more degrees of freedom. Section \ref{Sec_dyna_weight_search} discusses more details on the observability issue.

\subsubsection{IRLS for RKHS space}

To successfully implement RKHS formulation \eqref{GMM_object_func} on correspondence-free PnP, two problems should be solved: real-time optimization (over $10 Hz$) and approximation of maximization with minimization constraint. Most previous studies have focused on achieving high accuracy, often neglecting computational efficiency. Thus, derivative-free and easy-to-use solvers like the Powell or Nelder-Mead are used in these studies. Only \cite{groher2009deformable} adopts a 2nd-derivative-based solver BFGS for fast optimization. Additionally, maximization with Gaussian kernels is notoriously difficult. The Gaussian kernel cannot be written in a binomial form suitable for least squares (second-order derivative) solver. Consequently, CVO~\cite{clark2021nonparametric} uses 4th-order Taylor expansion to determine an optimal step for the first-order derivative-based gradient descent method. However, as a first-order derivative-based method, it only achieves 2-5 Hz on modern CPU, even with multi-core computation for typical point cloud registration problems.\par

IRLS algorithm~\cite{ray2024rkhsba} is applied to realize both real-time optimization (over $10 Hz$) and maximizing with minimization constraint. In each iteration, our IRLS uses \textit{Least Square} formulation \eqref{2Ddis_object_func_0_0} to search for the descending direction and employs RKHS values (in \eqref{GMM_object_func}) to iteratively adjust the searching direction. As demonstrated in the experiment by~\cite{ray2024rkhsba}, the second-order Least Square algorithm IRLS requires much less iterative searching than first-order searching in~\cite{clark2021nonparametric}. In the iterative optimization of \eqref{GMM_object_func}, the searching direction and step size of $\mathbf{T}^{k}$ in step $k$ are determined by minimizing the approximate equation as
\begin{equation}
\begin{aligned}
	\label{GMM_object_func_irls}
	&\mathbf{T}^{k+1}\!=\! \operatorname{min}_ {\mathbf{T},\theta} \sum_{i \in \Omega}\mathrm{w}_i \sum_{j \in \Delta}\mathrm{w}_{ij}^{k} \lVert\pi(\mathbf{T}\mathbf{p}_i,\mathbf{K})-\mathbf{q}^k_j\rVert^2_2 + \lambda \mathrm{E}_{init},
\end{aligned}
\end{equation}

\noindent where $\mathrm{w}_{ij}^{k} = \exp \left(-\frac{\lVert \pi(\mathbf{T}^{k}\mathbf{p}_i,\mathbf{K})-\mathbf{q}_j \rVert^2_2}{2\ell^2}\right)$ and fixed as the weight in the iterative searching. $\mathbf{T}^{k}$ is the state estimated in step $n$ and indifferentiable in $\mathrm{w}_{ij}^{n}$. $\mathbf{T}$ in \eqref{GMM_object_func_irls} is initialized with $\mathbf{T}^{k}$. \textbf{It should be emphasized the sum of Gaussian kernels $E_{data}$ is the only criteria to determine if $\mathbf{T}^{k+1}$ is accepted.} IRLS primarily facilitates the search for the descending direction. The direction and step size in \eqref{GMM_object_func_irls} are determined analytically in the least squares form for fast searching. \par 

The work of~\cite{ray2024rkhsba} proved that the IRLS form is the exact first-order derivative of the original function. Experiments in our previous work~\cite{song2023iterative} and \cite{ray2024rkhsba} demonstrate that second-order IRLS's version (approximate solver) of RKHS achieves similar performance as first-order gradient descent solver (exact solver). In the descending direction searching step, IRLS converts kernel sum maximization into reweighted minimization and thus can be combined with minimization constraint. However, \eqref{GMM_object_func_irls} is not the final solution. Instead of directly using \eqref{GMM_object_func_irls} in correspondence-free PnP (like ~\cite{song2023iterative}), the next section (Section \ref{Sec_dyna_weight_search}) reveals the observability issue and push \eqref{GMM_object_func_irls} toward a more robust algorithm.

\subsection{Dynamic weighted searching}
\label{Sec_dyna_weight_search}
In our preliminary 
work~\cite{song2023iterative}, we observe that ``point-wise registration is still successful even though the estimated pose is wrong'' and credit it as ``an extra amount of local minima'' in correspondence-free PnP. This section reveals that this problem is due to the numerical unobservability issue in correspondence-free PnP. Observability is defined as the ability to fully and uniquely recover the system state from a finite number of observations of its outputs and the knowledge of its controls~\cite{bar2004estimation}. 3D-2D registration suffers from two types of observability issues: infinitely small scale and rotation and translation ambiguity.\par

\begin{figure*}[!h]
    \centering
    \subfloat{
        \begin{minipage}[]{1\textwidth}
            \centering
            \includegraphics[width=1\linewidth]{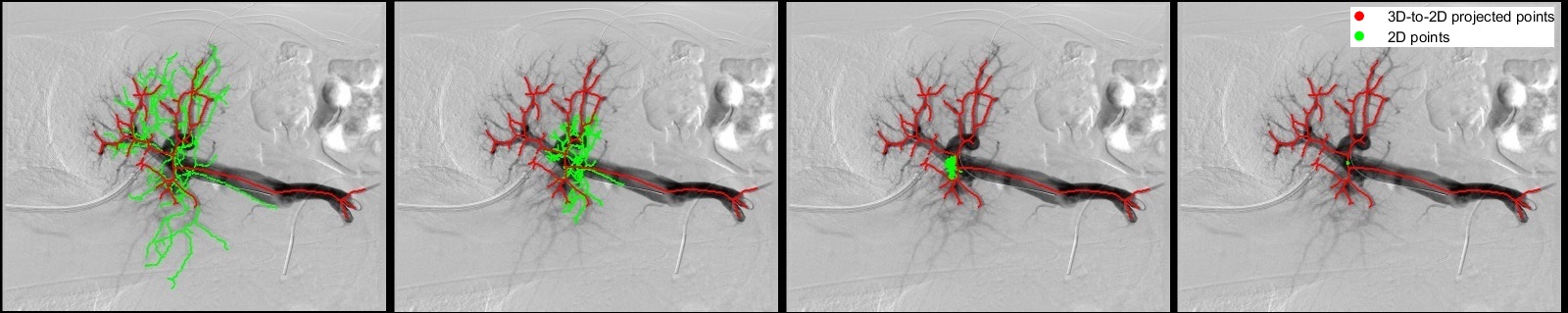}
        \end{minipage}
    }
    \caption{Illustrated is an example of scale ambiguity in the correspondence-free PnP process. The red and green points represent the 2D points and projected 3D points, respectively. As the 3D points move farther away from the sensor plane, their projected 2D points continue to shrink.}
    \label{fig_scale_ambiguity}
\end{figure*}

\subsubsection{A special case: Infinite small scale}
\label{Sec_a_special_case}
Unlike 3D point cloud registration scenarios, 3D-2D registration suffers from scale ambiguity, leading to a unique issue: infinitely small scale. With an infinitely large Z direction on $\mathbf{T}$, $\pi(\mathbf{T}\mathbf{p}_i,\mathbf{K})$ projects the entire group $\{\mathbf{p}\}$ on a single point regardless of noises or outliers as shown in Fig. \ref{fig_scale_ambiguity}. Consequently, both Euclidean loss in \eqref{2Ddis_object_func_0_0} (all distances are $0$) and RKHS kernel in \eqref{GMM_object_func} (all kernel values are maximum) achieves optimal in this infinite small scale. Thus, $E_{init}$ in \eqref{GMM_object_func} is involved to constrain the pose. Furthermore, the projection can be registered to any point in $\{\mathbf{q}\}$. \par 

Infinitely small scale is a special case that is strictly unobservable. Based on the equations, both L2 and RKHS losses indicate that this unwanted solution is optimal. Fortunately, it cannot always be reached since the $\mathrm{SE}(3)$ distance from the initial input to the infinitely small scale is large, and correspondence-free PnP is a nonconvex problem. When the 3D shape is correctly registered to the 2D shape, the system stops in the "local minimum" (correct but not optimal in L2 or RKHS loss).\par  

\subsubsection{Rotation and translation ambiguity in correspondence-free PnP}
\label{sec_rot_trans_ambiguity}

Another observability issue arises because point-wise correspondences cannot be obtained. As a result, rotation and translation are challenging to disentangle in practice, especially if the distance between the camera center and the object is large. Specifically, translation can be substituted by rotation. Define ${\mathbf{q}^R_i} \in \mathbb{R}^3$ and ${\mathbf{q}^T_i} \in \mathbb{R}^3$ as $\mathbf{p}_i$'s 2D projected points. The point ${\mathbf{q}^R_i}$ is transformed by rotation only, while the point set ${\mathbf{q}^T_i}$ is transformed by translation only. Their definitions are:

\begin{equation}
    \mathbf{q}^R_i = \pi(\operatorname{T}(\mathbf{R},0)\mathbf{p}_i,\mathbf{K}) = 
    \begin{bmatrix}
        \mathrm{f}_x\frac{\mathbf{R}_x\mathbf{p}_i}{\mathbf{R}_z\mathbf{p}_i}+\mathrm{c}_x\\
        \mathrm{f}_y\frac{\mathbf{R}_y\mathbf{p}_i}{\mathbf{R}_z\mathbf{p}_i}+\mathrm{c}_y\\

        1
    \end{bmatrix},
\end{equation}

\begin{equation}
    \mathbf{q}^T_i = \pi(\operatorname{T}(0,\mathbf{t})\mathbf{p}_i,\mathbf{K})=
    \begin{bmatrix}
        \mathrm{f}_x\frac{\mathbf{p}_i|_x+\mathbf{t}_x}{\mathbf{p}_i|_z+\mathbf{t}_z}+\mathrm{c}_x\\
        \mathrm{f}_y\frac{\mathbf{p}_i|_y+\mathbf{t}_y}{\mathbf{p}_i|_z+\mathbf{t}_z}+\mathrm{c}_y\\

        1
    \end{bmatrix},
\end{equation}

\noindent where $\operatorname{T}(\cdot,\cdot)$ convert the input rotation and translation into $\mathrm{SE}(3)$ transformation. $\{\mathbf{R},\mathbf{t}\}$ is one arbitrary rigid body transformation, $\mathbf{R}_x, \mathbf{R}_y, \mathbf{R}_z$ are row elements of rotation $\mathbf{R} \in \mathrm{SO}(3)$ and $\mathbf{t}_x, \mathbf{t}_y, \mathbf{t}_z$ are elements of translation $\mathbf{t} \in \mathbb{R}^3$. $\mathrm{f}_x$ and $\mathrm{f}_y$ are focal length of the camera and $\mathrm{c}_x,\mathrm{c}_y$ the optical center. \sjw{Take United Imaging's uAngio 960 as an example, the typical values of $\mathrm{f}_x$ and $\mathrm{f}_y$ (source to image distance multiply by pixel spacing for image size $512\times512$) and $\mathbf{p}_i|_z$ (source to object distance) are $1877$ and $800 mm$. $\mathbf{p}_i|_x$ and $\mathbf{p}_i|_y$ ranges from $-90 mm$ to $90 mm$.}\par

Define $\mathbf{p}_i|_x$, $\mathbf{p}_i|_y$ and $\mathbf{p}_i|_z$ as the x, y and z elements of $\mathbf{p}_i$. We show that with an arbitrary translation $\mathbf{t}$ and for a point cloud with $\mathbf{p}_i|_z \gg \mathbf{p}_i|_x$ and $\mathbf{p}_i|_z \gg \mathbf{p}_i|_y$, a small rotation $\mathbf{R}$ achieves similar transformation numerically. It should be noticed that we refer to \textbf{numerical ambiguity} with the existence of small noises and correspondences errors. If accurate observation and correspondences are known, rotation and translation are separable. We show that a small rotation $\mathbf{R}$, represented as local disturbance on Lie algebra, can bring similar observations in correspondence-free PnP. Applying the first order expansion of $\mathbf{R}$ on exponential map:

\begin{equation}
    \mathbf{R} \simeq \mathbf{I} + \phi^\wedge=
    \begin{bmatrix}
        1 & -\phi_3 & \phi_2\\
        \phi_3 & 1 & -\phi_1\\
        -\phi_2 & \phi_1 & 1
    \end{bmatrix},
\end{equation}

\noindent where all elements $\phi_1$, $\phi_2$ and $\phi_3$ in $\phi$ are \textbf{much smaller than 1}. Therefore, the projections of the rotation disturbance $\mathbf{R}$ is\par

\begin{equation}
\begin{aligned}
    \mathbf{q}^R_i &= \pi(\operatorname{T}(\mathbf{R},0)\mathbf{p}_i,\mathbf{K})\\ 
    &\simeq 
    \begin{bmatrix}
        \mathrm{f}_x\frac{\mathbf{p}_i|_x-\phi_3\mathbf{p}_i|_y+\phi_2\mathbf{p}_z}{-\phi_2\mathbf{p}_i|_x+\phi_1\mathbf{p}_i|_y+\mathbf{p}_i|_z}+\mathrm{c}_x\\
        \mathrm{f}_y\frac{\phi_3\mathbf{p}_i|_x+\mathbf{p}_i|_y-\phi_1\mathbf{p}_z}{-\phi_2\mathbf{p}_i|_x+\phi_1\mathbf{p}_i|_y+\mathbf{p}_i|_z}+\mathrm{c}_y\\

        1
    \end{bmatrix}.
\end{aligned}
\end{equation}

Take the first element of $\mathbf{q}^R_i|_x$ and $\mathbf{q}^T_i|_x$ as an example, its difference is:
\begin{equation}
\label{Eq_ambiguity}
 \begin{aligned}
   &\mathbf{q}^R_i|_x- \mathbf{q}^T_i|_x
   =\mathrm{f}_x\left(\frac{\mathbf{p}_i|_x-\phi_3\mathbf{p}_i|_y+\phi_2\mathbf{p}_i|_z}{-\phi_2\mathbf{p}_i|_x+\phi_1\mathbf{p}_i|_y+\mathbf{p}_i|_z} - \frac{\mathbf{p}_i|_x+\mathbf{t}_x}{\mathbf{p}_i|_z+\mathbf{t}_z}\right)\\
   &\stackrel{(\mathrm{i})}\simeq\mathrm{f}_x\left(\frac{\mathbf{p}_i|_x-\phi_3\mathbf{p}_i|_y+\phi_2\mathbf{p}_i|_z}{\mathbf{p}_i|_z} - \frac{\mathbf{p}_i|_x+\mathbf{t}_x}{\mathbf{p}_i|_z+\mathbf{t}_z}\right)\\
    &=\mathrm{f}_x\left(\phi_2-\phi_3\frac{\mathbf{p}_i|_y}{\mathbf{p}_i|_z}+\frac{\mathbf{p}_i|_x\mathbf{t}_z}{\mathbf{p}_i|_z(\mathbf{p}_i|_z+\mathbf{t}_z)}-\frac{\mathbf{t}_x}{(\mathbf{p}_i|_z+\mathbf{t}_z)}\right)\\
    &=\mathrm{f}_x\phi_2-\frac{\mathrm{f}_x\mathbf{t}_x}{(\mathbf{p}_i|_z+\mathbf{t}_z)}-\mathrm{f}_x\phi_3\frac{\mathbf{p}_i|_y}{\mathbf{p}_i|_z}+\frac{\mathrm{f}_x\mathbf{p}_i|_x\mathbf{t}_z}{\mathbf{p}_i|_z(\mathbf{p}_i|_z+\mathbf{t}_z)}\\
    &\stackrel{(\mathrm{ii})}\simeq\mathrm{f}_x\phi_2-\frac{\mathrm{f}_x\mathbf{t}_x}{(\mathbf{p}_i|_z+\mathbf{t}_z)}.\\
\end{aligned}   
\end{equation}

\noindent (i): $\mathbf{p}_i|_z \gg \mathbf{p}_i|_x$, $\mathbf{p}_i|_z \gg \mathbf{p}_i|_y$ and $\phi_i \ll 1$. (ii): $\left\vert \mathrm{f}_x\phi_3\frac{\mathbf{p}_i|_y}{\mathbf{p}_i|_z} \right\vert \ll 1$ \sjw{($\mathrm{f}_x$ is less than $3\mathbf{p}_i|_z$ in modern DSA devices)}, $\left\vert \frac{\mathrm{f}_x\mathbf{p}_i|_x\mathbf{t}_z}{\mathbf{p}_i|_z(\mathbf{p}_i|_z+\mathbf{t}_z)} \right\vert \ll 1$ and $\left\vert\frac{\mathrm{f}_x\mathbf{t}_x}{(\mathbf{p}_i|_z+\mathbf{t}_z)} \right\vert$ is close to 1. Therefore, $\phi_2=\frac{\mathbf{t}_x}{(\mathbf{p}_i|_z+\mathbf{t}_z)}$ ensures $\mathbf{q}^R_i|_x$ very close to $\mathbf{q}^T_i|_x$ given an arbitrary $\mathbf{t}$. Similarly, we can also conclude that $\phi_1=\frac{\mathbf{t}_y}{(\mathbf{p}_i|_z+\mathbf{t}_z)}$ makes $\mathbf{q}^R_i|_y$ close to $\mathbf{q}^T_i|_y$ given an arbitrary $\mathbf{t}$. $\phi = \left[\frac{\mathbf{t}_y}{\mathbf{p}_i|_z+\mathbf{t}_z},\frac{\mathbf{t}_x}{\mathbf{p}_i|_z+\mathbf{t}_z},\phi_3 \right]$ is a dual numerical solution to a chosen $\mathbf{t}$. Since a small disturbance on rotation can have a similar effect as a given translation, the effect of translation can be substituted by a rotation.\par

\begin{figure}[t]
    \centering
    \includegraphics[width=0.5\columnwidth]{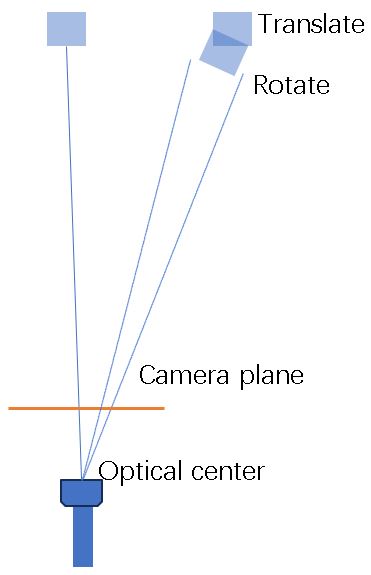}
    \caption{The figure shows the ambiguity of rotation and translation. Rotating and translating the cube object yield similar observations (the projected silhouette).}
    \label{fig_ambiguity_rot_trans}
\end{figure}

Fig.~\ref{fig_ambiguity_rot_trans} intuitively shows the ambiguity between rotation and translation of the object. It reveals that the curve of the circumcircle movement (rotation) and tangent movement (translation) produce similar projections. \sjw{Therefore, translation and rotation are not separable numerically if the correspondences are unknown.} Moreover, it indicates that since a much smaller rotation can substitute a given translation, the solver tends to optimize rotation in the $\mathrm{SE}(3)$ space regardless of the pose representation. Our previous work~\cite{song2023iterative} validates that the iterative searching procedure tends to update rotation more.\par  

\subsubsection{Alternative direction searching by dynamic re-weighting}

Section \ref{sec_rot_trans_ambiguity} indicates that the massive amount of local minima is due to the numerical ambiguity between rotation and translation. As a result, the authentic one-step joint estimation in correspondence-free PnP inevitably falls into a ``bad'' local minima where rotation predominates the entire optimization. The numerous local minima is the main obstacles for better registration. Previous works use particle filters algorithm~\cite{seo2013optimal,tian2022large} to reduce the impact of the local minimum. The aim of this work is to refine the registration by proposing an alternative direction searching approach to two problems: the original problem and an auxiliary problem. The auxiliary formulation is built by matching the reweighted (or subset for faster implementation) of source 3D points to target 2D points: 

\begin{equation}
	\label{2Ddis_node_func_IRLS}
 \begin{aligned}
&\mathbf{T}^*=\\
&\operatorname{argmax}_ {\mathbf{T}} \sum_{\mathbf{p}'_i \in \{\mathbf{p'}\}}\mathrm{w}'_i \sum_{\mathbf{q}^{k'}_j \in \{\mathbf{q'}\}}\exp \left(-\frac{||\pi(\mathbf{T}\mathbf{p}'_i,\mathbf{K})-\mathbf{q}^{k'}_j||^2_2}{2\ell^2}\right), 
 \end{aligned}
\end{equation}

\begin{equation}
	\label{2Ddis_node_func_IRLS1}	
 \begin{aligned}
 &\mathbf{T}^*=\\
 &\operatorname{argmax}_ {\mathbf{T}} \sum_{\mathbf{p}_i \in \{\mathbf{p}\}}\mathrm{w}_i \sum_{\mathbf{q}^{k}_j \in \{\mathbf{q}\}}\exp \left(-\frac{||\pi(\mathbf{T}\mathbf{p}_i,\mathbf{K})-\mathbf{q}^k_j||^2_2}{2\ell^2}\right),
 \end{aligned}
\end{equation}

\noindent where $\mathbf{p}'_i \in \{\mathbf{p}'\}$ ($\mathbf{p}'_i \in \mathbb{R}^4$) is a subset of $\{\mathbf{p}\}$ and $\mathbf{q}^{k'}_j \in \{\mathbf{q}'\}$ ($\mathbf{q}^{k'}_j \in \mathbb{R}^3$) is the corresponding closest 2D point. $\{\mathrm{w}'_i\}$ are the corresponding modified weights. Therefore, the proposed 'dynamic re-weighting' consists of reweighting the original \eqref{2Ddis_node_func_IRLS1} alongside simplifying the problem via subset selection (i.e., directly setting certain weights to 0), which reformulates an an auxiliary problem. Different from \eqref{2Ddis_node_func_IRLS1} using iterative correspondences searching, \textbf{\eqref{2Ddis_node_func_IRLS} fixed the correspondences between $\mathbf{p}_i'$ and $\mathbf{q}^{k'}_j$ in initialization step for faster convergence}. Algorithm \ref{Algorithm_ADMM} and Fig.~\ref{fig_framework} shows the procedure of the optimization. In each alternating step, the ``bad'' pose local minima provides correspondences of the vessel nodes.\par

\begin{figure}[!h]
    \centering
    \subfloat{
        \begin{minipage}[]{0.9\textwidth}
            \centering
            \includegraphics[width=1\linewidth]{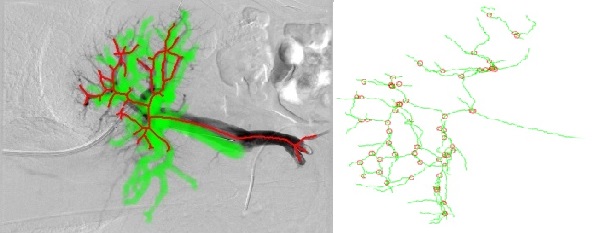}
        \end{minipage}
    }
    \caption{The figure visualizes the process of alternative searching. The left green tree is the projected 2D vessel. The right green tree is the original 3D vessel. The subproblem is set as matching the cross points to the 2D vessel centerline.}
    \label{fig_alternative_search_sample}
\end{figure}

Fig. \ref{fig_alternative_search_sample} illustrates a sample implementation of Algorithm \ref{Algorithm_ADMM} and Fig.~\ref{fig_framework}. The right green tree is the complete 2D points set $\{\mathbf{p}\}$, while the red nodes indicate the subset $\{\mathbf{p}'\}$, derived from $\{\mathbf{p}\}$. It should be noted that the subset $\{\mathbf{p}'\}$ can be selected through various methods, such as sub-branches, main branches, or individual nodes, as discussed in this article. This work sets all elements in $\{\mathrm{w}'_i\}$ and $\{\mathrm{w}'_i\}$ as $1$. For faster implementation, the sparse downsampled nodes' correspondences $\{\mathbf{q}'\}$ are fixed given the initial pose $\mathbf{T}^0$ due to the incremental searching. The selection process depends on the specific scenario. Then, Algorithm \ref{Algorithm_ADMM} can be implemented based on original 2D points $\{\mathbf{p}\}$ and the selected subset $\{\mathbf{p}'\}$.\par

\begin{figure}[t]
    \centering
    \includegraphics[width=1\columnwidth]{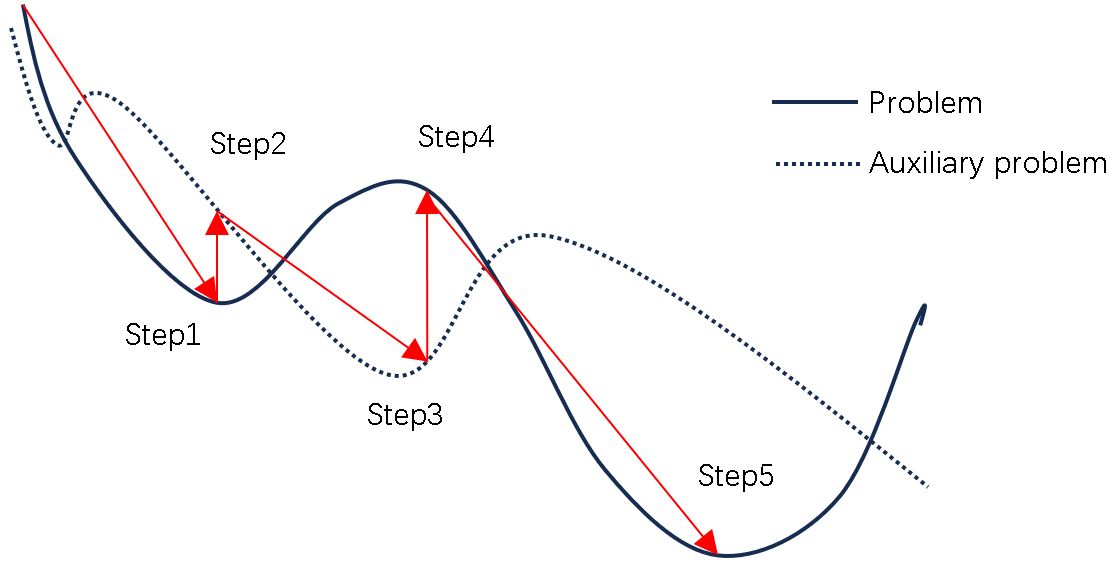}
    \caption{The figure illustrates the concept of alternative direction searching. After the state falls into a local minimum (steps 1 and 3), the auxiliary problem helps the state jump out (steps 2 and 4). This alternative direction searching method thus aids the algorithm in converging to a better local minimum.}
    \label{fig_alternative_search}
\end{figure}

Fig. \ref{fig_alternative_search} shows the basic idea of the proposed alternative searching. With the initial of the pose $\textbf{T}^0$, a 3D to 2D registration is initialized. Then, an exponential kernel-based PnP is applied alternatively to both full set $\{\mathbf{p}\}$ and subset $\{\mathbf{p}'\}$ shown in \eqref{2Ddis_node_func_IRLS} and \eqref{2Ddis_node_func_IRLS1}. \textbf{In the alternative searching, the estimated pose $\mathbf{T}^k$ is used to initialize registration for next step.}

A trick is used in the alternative searching displayed in Fig. \ref{fig_alternative_search}. Considering that translation and rotation are numerically ambiguous, we propose simplifying \eqref{GMM_object_func_irls} to a rotation-only formulation for faster convergence. It is formulated as

\begin{equation}
\begin{aligned}
	\label{GMM_object_func_rot_only}
	&\mathbf{R}^*=\\
 &\operatorname{argmax}_ {\mathbf{R}} \sum_{i \in \Omega(i)}\mathrm{w}_i \! \sum_{j \in \Delta}\exp \left(-\frac{||\pi(\operatorname{T}(\mathbf{R},0)\mathbf{p}_i,\mathbf{K})\!-\!\mathbf{q}_j||^2_2}{2\ell^2}\right) \\
\end{aligned}.
\end{equation}


Equation \eqref{GMM_object_func_rot_only} avoids the scale problem, converge faster than \eqref{GMM_object_func_irls} and does not have 2D scale ambiguity problem.\par 

\begin{algorithm}[t]
\small 
	\caption{Alternative searching.}
	\label{Algorithm_ADMM}
	\KwIn{$\{\mathbf{p}\}$,$\{\mathbf{q}\}$,$\mathbf{K}$}
	\KwOut{Optimal sensor pose $\mathbf{T}^*$}
    \nextnr
    $k=0$\\
	\nextnr
    Search $\mathbf{T}^k$ based on \eqref{GMM_object_func_rot_only} ($\mathbf{R}^k$ to be specific). \\
    \nextnr
    Build 3D-2D correspondences $\{\mathbf{p}_i,\mathbf{q}^k_i\}$, the median TRE $\xi_k$ of $\{\mathbf{p}\}$ and the median TRE $\xi'_k$ of $\{\mathbf{p}'\}$. \\
    \nextnr
	\Do{$\xi_k < \xi$}
    {
        \nextnr
        Search $\mathbf{T}^k$ with the registrations $\{\mathbf{p}_i,\mathbf{q}^{k}_i\}$ and \eqref{2Ddis_node_func_IRLS}.\\ 
        \nextnr
        Build registrations $\{\mathbf{p}'_i,\mathbf{q}^{k'}_i\}$ with $\mathbf{T}^k$.\\
		\nextnr
        Search $\mathbf{T}^k$ with the registrations $\{\mathbf{p}'_i,\mathbf{q}^{k'}_i\}$ and \eqref{2Ddis_node_func_IRLS1}.\\
        \nextnr
        Build registrations $\{\mathbf{p}_i,\mathbf{q}^{k}_i\}$ with $\mathbf{T}^k$.\\
        \nextnr
        Estimate median TRE $\xi_k$ from $\{\mathbf{p}_i,\mathbf{q}^{k}_i\}$.\\
		\nextnr
		$k=k+1$\\
    }
	Notation: The scalar $\xi$ is the threshold. Median Target Registration Error (TRE) measures the median TRE and TRE denote the Euclidian distances between the reprojected 3D vessel and 2D vessel.\\
\end{algorithm}

All rigid transformation $\mathbf{T}$ are incrementally optimized on the $\mathrm{SE}(3)$ manifold~\cite{boumal2023introduction}, in contrast to the Euler angles and $\mathbb{R}^3$ translation used in existing research. Optimization on the continuous $\mathrm{SE}(3)$ manifold avoids gimbal lock degeneration in the Euler angles space by following geodesics along the manifold for searching. Levenberg-Marquardt algorithm is adopted as the minimizer.\par

\subsection{Differences from early works}

It should be noticed that the rotation-translation ambiguity analyzed in this work differs fundamentally from the classical finite-fold ambiguity in multi-view geometry~\cite{triggs2000bundle,hartley2003multiple}. The classical ambiguity, a well-known phenomenon arising from essential matrix decomposition with \textbf{known correspondences}, presents a discrete choice among a few geometrically valid poses. In contrast, the ambiguity inherent to \textbf{correspondence-free} PnP is a distinct, optimization-induced coupling: \textbf{in the absence of known correspondences}, rotation and translation parameters become numerically entangled during iterative alignment, allowing one to erroneously compensate for the other and leading to a continuum of spurious local minima. This continuous, numerically pathological landscape—where point-wise matching appears successful despite an incorrect global pose—is the core challenge addressed in our formal analysis (Eq. 10) and is directly mitigated by the proposed dynamic re-weighting strategy.

The proposed alternating optimization shares the iterative alternating structure with Expectation Maximization (EM), yet with different mechanism. EM addresses latent variables by alternating between probabilistic inference (E-step) and model parameter update (M-step). In contrast, our method alternates between solving the full registration problem and a deliberately simplified sub-problem to provide complementary search directions. This deterministic strategy leverages geometric priors to escape local minima, differing from particle filters that rely on stochastic sampling. 
	
	\section{Results and discussion}
	\label{section_result}

    The \textbf{numerical ambiguity} issue raised from \eqref{Eq_ambiguity} and Fig.~\ref{fig_ambiguity_rot_trans} was first validated with a toy model simulation. Then, synthetic, in-vivo, and ex-vivo experiments using pre-operative 3D Computed Tomography Angiograms (3D-CTA) and 2D fluoroscopic X-rays were performed to validate the accuracy, efficiency, and performance of the DynaWeightPnP. Finally, an ablation study was conducted to reveal the efficiencies of the proposed modules. As the task of 6DoF texture-less objects pose tracking normally has clear shape and no partial overlap, this research uses 3D-2D vessel registration for validation.

    \begin{figure}[t]
    \centering
    \includegraphics[width=1\columnwidth]{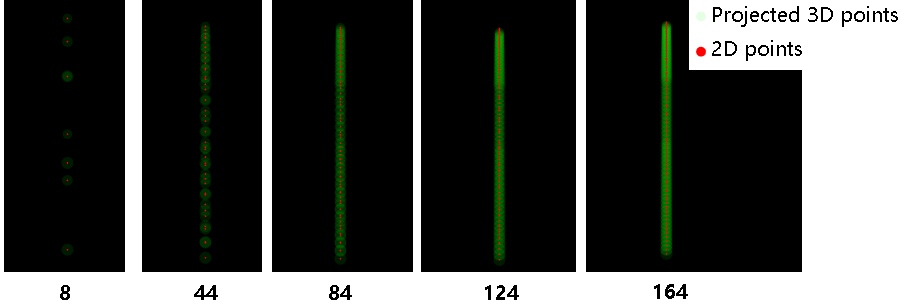}
        \caption{The figure shows the points’ density in the square toy model. The numbers are the discretized points in the square. Green points are the projected 3D points, and red points are the 2D points.}
        \label{fig_synth_results}
    \end{figure}

    \begin{table*}[]
    \centering
    \caption{The table reports the average TRE, angle differences (Angle in degree), and translational differences (Dist.) observed in tests using the square toy model. The number of points in the discretized square ranges from 8 to 164. Angle is measured in degrees, TRE in pixels, and Dist. in a uniform unit of pose.}
    \resizebox{\textwidth}{!}{
    \begin{tabular}{llllllllllllllll}
    \toprule
      Pts Num    & \multicolumn{3}{l}{8} & \multicolumn{3}{l}{44} & \multicolumn{3}{l}{84} & \multicolumn{3}{l}{124} & \multicolumn{3}{l}{164} \\ \midrule
                  & TRE      & Angle    & Dist.   & TRE       & Angle    & Dist.    & TRE       & Angle    & Dist.    & TRE       & Angle    & Dist.     & TRE       & Angle     & Dist.    \\ \midrule
    DT-ICP        & 0.000    & 0.000    & 0.000   & 0.000     & 0.000    & 0.000    & 2.226     & 0.862    & 0.332    & 4.619     & 0.243    & 0.108     & 4.738     & 0.136     & 0.137    \\
    NICP          & 0.000    & 0.000    & 0.000   & 0.795     & 0.311    & 0.099    & 0.742     & 0.292    & 0.099    & 6.650     & 0.214    & 0.100    & 5.193     & 0.147     & 0.100    \\
    DynaWeightPnP$^*$ & 0.000    & 0.000    & 0.000   & 0.018     & 0.012    & 0.003    & 0.024     & 0.016    & 0.004    & 3.234     & 1.379    & 0.383     & 2.402     & 0.989     & 0.284    \\ 
    \bottomrule
    \end{tabular}
    }
    \label{Table_toy_model_result}
    \end{table*}
	
	\subsection{Experiment setup}
    \subsubsection{data sets}
    Experiments include a toy model simulation, qualitative tests, and an ablation study. The toy model simulation strictly followed the setting in Fig. \ref{fig_ambiguity_rot_trans}, as it was implemented to validate our claim regarding the \textbf{numerical ambiguity} between rotation and translation. Qualitative tests were conducted on medical data sets, which consist of a synthetic data set, an in-vivo data set, and an ex-vivo data set\footnote{We strongly recommend readers watch the attached video for the experiments.}. \par

    The chosen medical data includes pre-operative 3D-CTA and intra-operative 2D fluoroscopic X-ray images, which are ideal for testing multi-modal 3D-2D alignment. The simulation data set was generated by applying a random $\mathrm{SE}(3)$ transformation and 3D-to-2D projection to the segmented 3D vessel centerlines. All algorithms aligned these projected vessel centerlines with their original 3D shapes by estimating a rigid transformation. As highlighted by~\cite{rivest2012nonrigid}, testing on simulations evaluates algorithm performance under ideal conditions with known sensor poses and without noise. Additionally, \textit{in-vivo} data sets from six patients were included: P1 (44), P2 (60), P3 (39), P4 (53), P5 (45), and P6 (37), with the number in parentheses indicating the image count.\par 
    
    An \textit{ex-vivo} data set was also generated by scanning a phantom using United Imaging's uAngio 960 device. 3D vessel centerlines were collected from pre-operative CTA images through segmentation, followed by online X-ray images.

    \subsubsection{Comparisons}

    DynaWeightPnP was compared against prior-free approaches AutoMask~\cite{steininger2012auto}, DT-ICP~\cite{rivest2012nonrigid}, RGBR~\cite{markelj2008robust}, Normalized ICP (NICP)~\cite{aylward2003registration} and our previous work which has no alternative searching (named Iterative PnP in previous work)~\cite{song2023iterative}. For simplicity, this article renames DynaWeightPnP without alternative searching~\cite{song2023iterative} as DynaWeightPnP$^*$. The original DT-ICP~\cite{rivest2012nonrigid} used BFGS as the second-order solver, while our implementation adopted the Levenberg-Marquardt solver on the Lie manifold for better convergence and faster speed. Our experiments show faster speed and less iteration than BFGS. We acknowledged the complexity and challenges associated with RL-related methods~\cite{miao2018dilated, miao2019agent} and were unable to implement them. As pointed out by Meng et al.~\cite{meng2022weakly}, these complex DNN-based approaches suffer from "the hardness of obtaining the ground-truth transformation parameters" and "real intra-operative DSAs can be incomplete." However, the failure to re-implement RL-based methods does not affect our major claim. \sjw{A deep learning comparison experiment, which may have a potentially inaccurate implementation, has been provided in the supplementary material.}\par 

    \subsubsection{Hardware and software}

    A commercial laptop, the ALIENWARE M17 R4 equipped with an Intel i7-10870H processor and 32GB of RAM, was utilized for the experiments. All prior-free methods were implemented on CPU end. Since no open-source implementations were available, we developed all approaches by ourselves.

    All previous works were re-implemented strictly following their descriptions. AutoMask was implemented in Python 3. Its derivative-free optimizer, Nelder-Mead, which fails in real-time performance. RGBR was implemented in Matlab, although its execution time was deemed negligible. DT-ICP, NICP, DynaWeightPnP$^*$, and DynaWeightPnP were implemented in C++ and integrated into the Robot Operating System (ROS) as a package~\cite{quigley2009ros}. Throughout the experiments, both 3D and 2D vessel data sets consisted of sizes ranging from 1500 to 3000 points. The hyperparameters $\lambda$ is set to $100$.

    \begin{figure*}[!h]
		\centering
		\subfloat{
			\begin{minipage}[]{1\textwidth}
				\centering
				\includegraphics[width=1\linewidth]{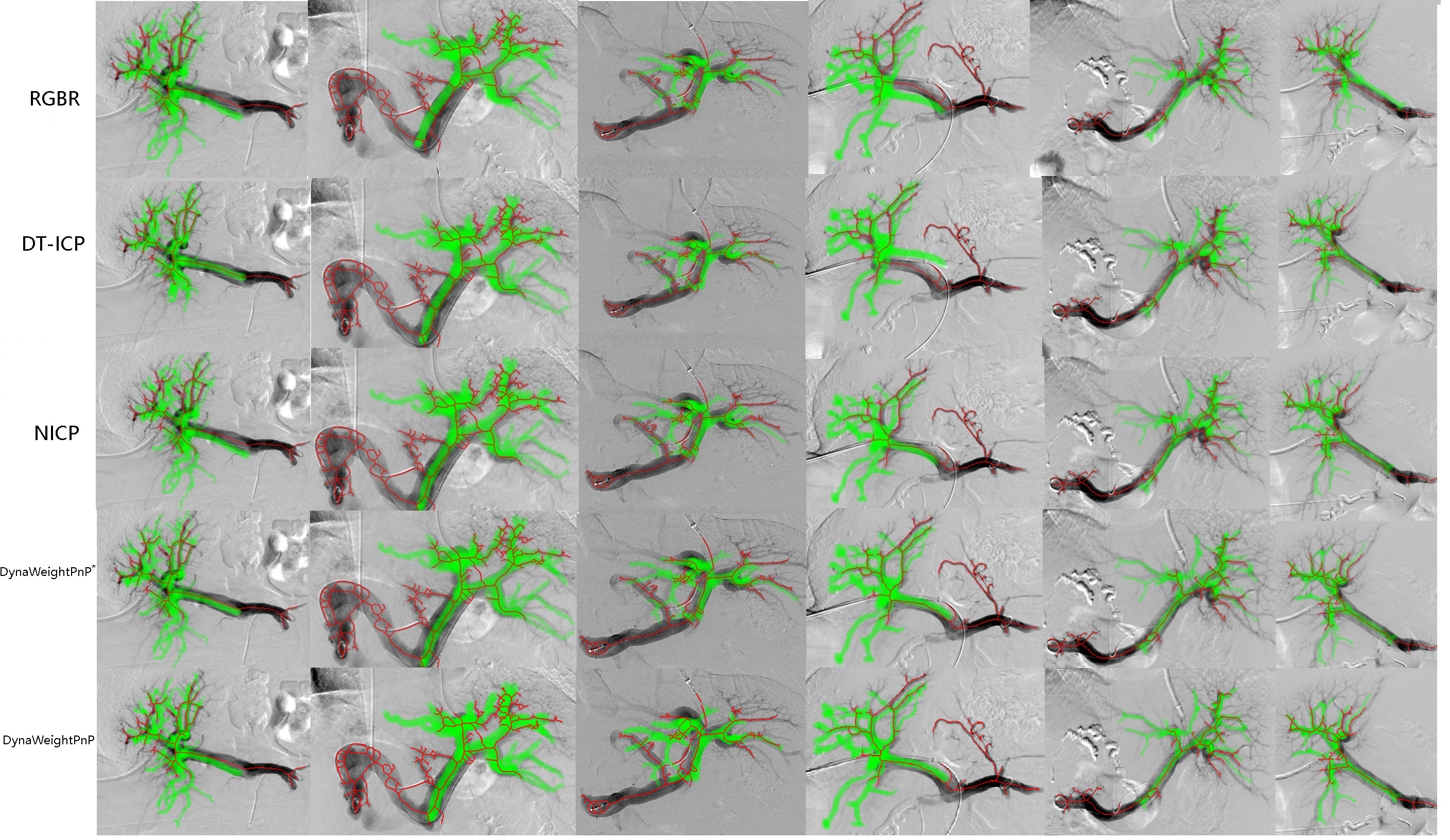}
			\end{minipage}
		}
		\caption{The figure illustrates alignment comparisons among RGBR, DT-ICP, NICP, DynaWeightPnP$^*$, and DynaWeightPnP. The green shape represents the projection of the pre-operative 3D vessel, while the red shape denotes the vessel's 2D centerline. Each of the six columns represents sample results from six patients.}
		\label{fig_invivo_results}
	\end{figure*}

    \begin{figure}[t]
    \centering
    \includegraphics[width=1\columnwidth]{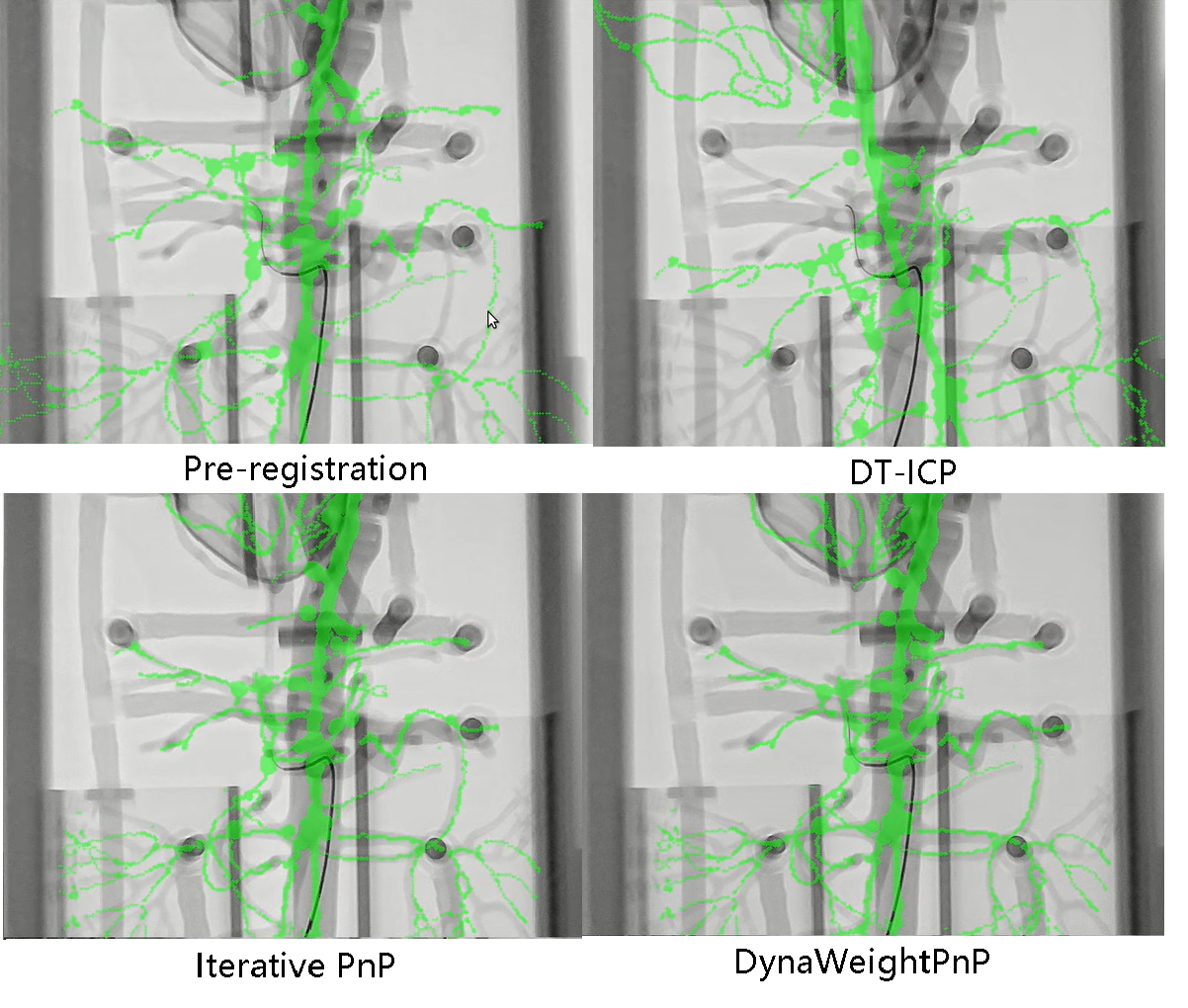}
        \caption{Illustrated is the alignment comparisons of DT-ICP, DynaWeightPnP$^*$ and DynaWeightPnP on an ex-vivo data sets.}
        \label{fig_exvivo_compare_results}
    \end{figure}
    

    \begin{table}[t]
\centering
\caption{The table shows the accuracy of the \textit{in-vivo} data sets. Median PR is the median of PR of all source points. GFR refers PR greater than $5 mm$.}
\begin{tabular}{lrllll}
\toprule
\multicolumn{1}{c}{\multirow{2}{*}{Method}} & \multicolumn{1}{c}{\multirow{2}{*}{GFR}} & \multicolumn{1}{c}{\multirow{2}{*}{\begin{tabular}[c]{@{}c@{}}Median PR\\ (mm)\end{tabular}}} & \multicolumn{2}{c}{Percentile (mm)} & \multicolumn{1}{c}{\multirow{2}{*}{\begin{tabular}[c]{@{}c@{}}Run time\\ (ms)\end{tabular}}} \\ \cline{4-5}
\multicolumn{1}{c}{}                        & \multicolumn{1}{c}{}                     & \multicolumn{1}{c}{}                                                                       & 95\%             & 75\%             & \multicolumn{1}{c}{}                                                                         \\ 
\midrule
AutoMask                              &   $100\%$                                         &  \multicolumn{1}{c}{17.01}                                                                                         &        29.75$^*$           &      23.26            & 150302.2                                                                                            \\
RGBR                                        &      $100\%$                                    & \multicolumn{1}{c}{5.99}                                                                   & 38.85            & 12.16            & -                                                                                            \\
DT-ICP                                      & \multicolumn{1}{l}{$21\%$}                                     & \multicolumn{1}{c}{4.84}                                                                                       & 50.04            & 14.11            & 13.2                                                                                         \\
NICP                                        & \multicolumn{1}{l}{$56\%$}                 & \multicolumn{1}{c}{\textbf{3.49}}                                                                                       & 37.09            & \textbf{10.52}            & 385.1                                                                                        \\
DynaWeightPnP$^*$                               & \multicolumn{1}{l}{$15\%$}                 & \multicolumn{1}{c}{6.17}                                                                                       & 38.89            & 13.72            & \textbf{16.9}                                                                                         \\
DynaWeightPnP                               & \multicolumn{1}{l}{$\mathbf{12\%}$}                 & \multicolumn{1}{c}{4.02}                                                                                       & \textbf{31.85}            & 11.35            & 33.2                                                                                        \\
\bottomrule
\end{tabular}
\label{Table_invivo_result}
\end{table}

\begin{table}[t]
\centering
\caption{\sjw{Wilcoxon signed-rank test results comparing DynaWeightPnP with other methods on \textit{in-vivo} data. Top: test based on 95\% Percentile (mm); Bottom: test based on Median PR (mm). Significance levels: *p<0.05, **p<0.01, ***p<0.001, n.s.: not significant.}}
\label{Table_wilcoxon_combined}

\smallskip
\noindent
\textbf{(a) Based on 95\% Percentile} \\
\begin{tabular}{lccccc}
\toprule
Method & W Statistic & p-value & Significance & Effect Size (r) \\
\midrule
DT-ICP & 0 & 0.016 & * & 0.879 \\
NICP & 0 & 0.016 & * & 0.879 \\
DynaWeightPnP$^*$ & 1 & 0.031 & * & 0.760 \\
\bottomrule
\end{tabular}

\vspace{0.3cm}

\noindent
\textbf{(b) Based on Median PR} \\
\begin{tabular}{lccccc}
\toprule
Method & W Statistic & p-value & Significance & Effect Size (r) \\
\midrule
DT-ICP & 1 & 0.031 & * & 0.760 \\
NICP & 9 & 0.422 & n.s. & 0.080 \\
DynaWeightPnP$^*$ & 12 & 0.656 & n.s. & -0.164 \\
\bottomrule
\end{tabular}

\vspace{0.1cm}
\footnotesize{\par\sjw{\textit{Note:} Wilcoxon signed-rank test (one-tailed) was performed for each method pair, 
testing the alternative hypothesis that DynaWeightPnP produces lower PR values than the comparison method. 
Effect size r was calculated as $r = Z/\sqrt{N}$ ($N=6$ and Z is the standardized test statistic from the standard normal distribution, derived from the estimated p-value.), where small ($|r| \geq 0.1$), medium ($|r| \geq 0.3$), and large ($|r| \geq 0.5$) effects are indicated.}}
\label{Table_wilcoxon_test}
\end{table}

	\subsubsection{Metric selection}

    This work follows previous research by adopting the widely used metrics for algorithm evaluation; metrics include Projection Residual (PR), Gross Failure Rate (GFR), computational time consumption, and pose difference (angular and translational). PR quantifies the Root Mean Square Error of all projected source points.

    \begin{equation}	
    \operatorname{PR}(\mathbf{T})=\sum_{i \in \Omega}||\pi(\mathbf{T}\mathbf{p}_i,\mathbf{K})-\overline{\mathbf{p}}_i||^2_2,
    \end{equation}

    \noindent where $\overline{\mathbf{p}}_i$ is the corresponding 2D point for $\mathbf{p}_i$ in simulation experiments or closest 2D points if data association is unknown. While PR may not effectively assess outliers, GFR identifies cases where PR exceeds a specified threshold. Median, 75th percentile, and 95th percentile PR values are reported as metrics. Moreover, considering this research only focuses on the 3D-2D registration algorithm, the time consumption of the image-to-shape procedure is not counted.

    \subsection{A toy model for rotation and translation ambiguity}

    This work simulated a toy model to validate the \textbf{numerical ambiguity} in correspondence-free PnP. It should be noted that \textbf{numerical ambiguity} is a general issue in correspondence-free PnP, independent of specific algorithms. The setup strictly followed the configuration in Fig. \ref{fig_ambiguity_rot_trans}. The virtual 2D image has dimensions of $1024 \times 1024$, and the optical center of the virtual camera is located at $(512,512)$. The virtual camera has a focal length of 520 on both axes. The square object is defined by its four corners $(5,0,10)$, $(10,0,10)$, $(10,0,15)$, and $(5,0,15)$ in counterclockwise order. A small disturbance with a radius of 1 was applied to the pose. Intuitively, with only four corners, four center points, and a small disturbance, the problem simplifies to a traditional PnP scenario where correspondences are known. Thus, we simulated the performance with varying points on the edge.\par 

    Table \ref{Table_toy_model_result} presents the results from three typical correspondence-free PnP algorithms, demonstrating the numerical ambiguity arising from ambiguous correspondences. The ambiguity in correspondences increases with the number of discretized points, affecting both pose estimation and TRE accuracy. This observation supports our claim from \cite{song2023iterative} that "point-wise registration can still be successful even when the estimated pose is inaccurate." Despite significant deviations of some poses from the reference poses, their TREs remain small. Fig.~\ref{fig_synth_results} shows sample 3D-2D registrations. The correct registration for 8 points can be easily retrieved, while it is difficult for 164 points. In summary, all correspondence-free PnP methods suffer from the local minima issue due to numerical ambiguity.\par


	\subsection{Tests on the synthetic and real-world data set}
	\label{section_in_vivo_ex_vivo}
    The proposed method mainly focuses on aligning pre-operative 3D-CTA and intra-operative 2D fluoroscopic X-ray images, as these multi-modal images are typically unified into vessels for registration and are well-suited for testing correspondence PnP methods. According to~\cite{pore2023autonomous}, endovascular interventions require "a precise understanding of the 3-D anatomy projected onto a 2-D image plane." The image-to-shape extraction procedure, which depends on the chosen scenario, falls outside the scope of this article. For 2D vessel segmentation, Otsu's method~\cite{otsu1979threshold} was employed, achieving approximately 400 ms processing time for an image size of $512 \times 512$ pixels. For 3D vessel segmentation, a DNN-based method provided by commercial software from United Imaging of Health Co., Ltd. was utilized. Other DNN-based approaches, such as those proposed by \cite{li2022dual, moccia2018blood}, are also applicable for 3D/2D image segmentation.

\begin{table}[]
\centering
\caption{The table reports the average pose of the simulation data by enforcing noise with angular standard deviation $2^\circ$ and translation standard deviation $5 mm$. GFR refers meanPR greater than $5$ pixels.}
\begin{tabular}{llllll}
\toprule
\multicolumn{1}{c}{\multirow{2}{*}{Method}} & \multicolumn{1}{c}{\multirow{2}{*}{GFR}} & \multicolumn{1}{c}{\multirow{2}{*}{\begin{tabular}[c]{@{}c@{}}MeanPR\\ (mm)\end{tabular}}} & \multicolumn{2}{c}{Percentile (mm)} & \multicolumn{1}{c}{\multirow{2}{*}{\begin{tabular}[c]{@{}c@{}}Run time\\ (ms)\end{tabular}}} \\ \cline{4-5}
\multicolumn{1}{c}{}                        & \multicolumn{1}{c}{}                     & \multicolumn{1}{c}{}                                                                     & 95\%             & 75\%             & \multicolumn{1}{c}{}                                                                         \\ 
\midrule
AutoMask                                & $0\%$                     &  0.85                                                                                         &   1.50               &    1.06              & 97004.0                                                                                                                                          \\
RGBR                                        & \multicolumn{1}{r}{$22\%$}                     & \multicolumn{1}{l}{2.21}                                                                     &      3.92       &    1.71              & -                                                                                            \\
DT-ICP                                   & $54\%$                 & 1.80                                                                                   & 6.65           & 3.49           & 13.2                                                                                         \\
NICP                                        & $8\%$                                     & 0.50                                                                                   & 3.11           & 1.14           & 385.0                                                                                        \\
DynaWeightPnP$^*$                                & $4\%$                                     & 0.39                                                                                   & 0.84           & 0.56           & \textbf{16.9}                                                                                         \\
DynaWeightPnP                           & $1\%$                                     & \textbf{0.24}                                                                                    & \textbf{0.50}            & \textbf{0.34}            & 31.5                                                                                         \\ 
\bottomrule
\end{tabular}
\label{Table_simu_result}
\end{table}

    \subsubsection{Simulation results}
    \label{section_simulation_results}
    The simulation experiment, conducted with simulated ground truth poses, initially sampled rigid poses 100 times with an angular standard deviation of $2^\circ$ and translation standard deviation of $5 mm$. \sjw{In the experiment, $\mathbf{T}_0$ is initialized with the perturbed pose.} These simulation experiments ensure a one-to-one correspondence without noise. Table \ref{Table_simu_result} demonstrates that the proposed DynaWeightPnP achieves superior performance compared to existing approaches. Compared with NICP, it reduces mean PR, $95\%$ TRE, and $75\%$ TRE by $50\%$, $83\%$, and $70\%$, respectively. Despite requiring additional time for alternative searching, it achieves a processing frequency of up to $31$ Hz.

    \subsubsection{In-vivo experiments}
    In-vivo experiments were conducted using data sets from patients P1 to P6. Unlike the simulation experiment in last section, the in-vivo experiments brings heavy amount of noises and outliers. Outliers are mainly resulted from some small vessels not extracted in one domain. Thus, the in-vivo experiment test the both accuracy and robustness of the chosen algorithms. Fig. \ref{fig_invivo_results} visually compares the performances of RGBR, DT-ICP, NICP, DynaWeightPnP$^*$, and DynaWeightPnP. Table \ref{Table_invivo_result} presents corresponding qualitative results. Table \ref{Table_wilcoxon_test} presents Wilcoxon signed-rank test to compare DynaWeightPnP with other methods. Wilcoxon signed-rank test results show that DynaWeightPnP achieves significantly lower projection residuals (PR) compared to DT-ICP (p=0.016, r=0.879), NICP (p=0.016, r=0.879), and DynaWeightPnP$^*$ (p=0.031, r=0.760), with large effect sizes ($r > 0.5$) in $95\%$ Percentile, confirming the superiority of DynaWeightPnP. It also can be observed from Fig. \ref{fig_invivo_results} that DynaWeightPnP$^*$ and DynaWeightPnP achieve similar registration efficacy and outperform comparing to other approaches. Upon closer inspection, DynaWeightPnP exhibits slightly improved registration accuracy in P3, P5, and P6 compared to DynaWeightPnP$^*$. Readers may notice that NICP achieves slightly better accuracy in terms of Median PR and $75\%$ Percentile PR. However, Fig. \ref{fig_invivo_results} shows that it sacrifices the accuracy of some branches in exchange for better registrations on the rest branches. That is why $56\%$ points of NICP exceeds $5 mm$. \par
    
    The enhanced global solution searching capability in DynaWeightPnP likely contributes to its superior accuracy. Readers are encouraged to refer to the supplementary material for the complete sequential registration results for the 6 \textit{in-vivo} data sets. These sequential results highlight DynaWeightPnP's global searching capability, as initial registrations notably deviate from the actual vessels.
    

    \subsubsection{Ex-vivo experiments}
    Ex-vivo experiments were conducted to validate the effectiveness of the employed RKHS loss. Fig. \ref{fig_big_to_small} demonstrates that the proposed RKHS loss effectively addresses partial overlap issues, whereas DT-ICP struggles with alignment. Fig. \ref{fig_exvivo_compare_results} further supports this finding, showing that DT-ICP fails to handle outliers (3D points without 2D correspondences) when the pre-operative 3D vessel is significantly larger than the 2D vessel. More qualitative results can be found in the supplementary material.

	\subsection{Time consumption}

    Table \ref{Table_simu_result} and Table \ref{Table_invivo_result} indicate that DT-ICP achieves the best performance at 75 Hz, while DynaWeightPnP$^*$ and DynaWeightPnP obtain 60 Hz and 31 Hz, respectively. It should be noted that the original DT-ICP~\cite{rivest2012nonrigid} reports a frequency of around 2 Hz. Our implementation, which utilizes the Levenberg–Marquardt solver on Lie manifold, requires significantly fewer iterations than their BFGS approach. Other contributing factors include differences in computational devices and data sets. Although the proposed DynaWeightPnP$^*$ is approximately $25\%$ slower than DT-ICP, it achieves higher accuracy and better robustness. Similarly, DynaWeightPnP's alternative searching approach doubles the computational burden. Unfortunately, the alternative searching procedure cannot be easily accelerated with parallel computation techniques. In summary, DynaWeightPnP$^*$ at 60 Hz excels in faster convergence and good registration accuracy, while DynaWeightPnP at 31 Hz achieves much better pose searching and registration accuracy.

	\subsection{Robustness to local minima}

    This work validates our main claim that "DynaWeightPnP reduces the impact of local minima" by appending the proposed alternative searching algorithm to different backbone methods. We explicitly test its ability to improve the performance of NICP and DynaWeightPnP$^*$ on simulation data sets. To achieve this, random disturbances were applied to the selected six in-vivo data sets. The variances for position and angle are ($5 mm$, $2^\circ$), ($8 mm$, $3^\circ$), and ($10 mm$, $5^\circ$). One experiment uses full 2D projections of the vessels, while the other randomly prunes five branches on both 3D and 2D data.

    Tables \ref{Table_IRLS_ADMM_nonoise} and \ref{Table_IRLS_ADMM_noise} record the pose and registration accuracy (in median TRE) on the simulation data sets. 10 trials are performed for each experiment. \sjw{In all experiments, $\mathbf{T}_0$ is initialized with the perturbed pose.} The results reveal several interesting phenomena. Firstly, NICP and DynaWeightPnP$^*$ only adjust rotation in their optimization procedures, validating our analysis that "smaller rotation is equivalent to the given translation". The distance defined on the Lie algebra $\mathrm{se}(3)$ prioritizes rotation during the search process. Secondly, the errors of the pose have been reduced by $20\%$ to $70\%$. Qualitative results show that the alternative searching reduced the average pose errors for NICP from ($5 mm$, $0.40^\circ$), ($8 mm$, $0.63^\circ$), ($10 mm$, $0.90^\circ$) to ($3.25 mm$, $0.20^\circ$), ($3.73 mm$, $0.21^\circ$), ($4.68 mm$, $0.28^\circ$) and for DynaWeightPnP$^*$ from ($5 mm$, $0.36^\circ$), ($8 mm$, $0.57^\circ$), ($10 mm$, $0.74^\circ$) to ($3.33 mm$, $0.20^\circ$), ($3.77 mm$, $0.21^\circ$), ($4.53 mm$, $0.28^\circ$) in Table \ref{Table_IRLS_ADMM_nonoise}. Similarly, the alternative searching reduced the average pose errors for NICP from ($5 mm$, $0.41^\circ$), ($8 mm$, $0.60^\circ$), ($10 mm$, $0.92^\circ$) to ($4.68 mm$, $0.31^\circ$), ($5.41 mm$, $0.35^\circ$), ($4.75 mm$, $0.29^\circ$) and for DynaWeightPnP$^*$ from ($5 mm$, $0.36^\circ$), ($8 mm$, $0.57^\circ$), ($10 mm$, $0.74^\circ$) to ($4.57 mm$, $0.29^\circ$), ($5.51 mm$, $0.34^\circ$), ($4.54 mm$, $0.28^\circ$) in Table \ref{Table_IRLS_ADMM_noise}. In summary, the errors of the pose have been reduced by $20\%$ to $70\%$.
    
	\begin{table*}[]
\caption{The table presents the results of NICP, NICP$^*$, DynaWeightPnP$^*$ and DynaWeightPnP. NICP$^*$ denotes NICP with alternative searching. Then, their results are adopted to initialize the Algorithm 1 for post refinement. The simulation 2D projections are generated by enforcing 10 disturbances on the original pose. \textbf{The simulated 2D vessel projection does not have any pruning.} Then, median TRE (``Median'') and pose difference (``Dist.'' for translational difference and ``Angle'' for angular difference) are calculated for the 10 trials. All results are presented in average.}
\setlength{\tabcolsep}{1pt}
\resizebox{\columnwidth}{!}{
\begin{tabular}{llrrrrrrrrrrrrrrrrrr}
\toprule
\multirow{2}{*}{}                                & \multirow{2}{*}{Algorithm} & \multicolumn{3}{c}{P\_1}                                                        & \multicolumn{3}{c}{P\_2}                                                       & \multicolumn{3}{c}{P\_3}                                                       & \multicolumn{3}{c}{P\_4}                                                       & \multicolumn{3}{c}{P\_5}                                                        & \multicolumn{3}{c}{P\_6}                                                        \\ \cline{3-20} 
                                                 &                            & \multicolumn{1}{c}{Median} & \multicolumn{1}{c}{Angle} & \multicolumn{1}{c}{Dist} & \multicolumn{1}{c}{Median} & \multicolumn{1}{c}{Angle} & \multicolumn{1}{c}{Dist} & \multicolumn{1}{c}{Median} & \multicolumn{1}{c}{Angle} & \multicolumn{1}{c}{Dist} & \multicolumn{1}{c}{Median} & \multicolumn{1}{c}{Angle} & \multicolumn{1}{c}{Dist} & \multicolumn{1}{c}{Median} & \multicolumn{1}{c}{Angle} & \multicolumn{1}{c}{Dist} & \multicolumn{1}{c}{Median} & \multicolumn{1}{c}{Angle} & \multicolumn{1}{c}{Dist} \\ \midrule
\multirow{4}{*}{\makecell[l]{$5mm$,\\$2^\circ$}} & NICP                       & 0.62                       & 0.50                      & 5.00                     & 0.60                       & 0.43                      & 5.00                     & 0.61                       & 0.39                      & 5.00                     & 0.55                       & 0.38                      & 5.00                     & 0.55                       & 0.37                      & 5.00                     & 0.54                       & 0.37                      & 5.00                     \\
                                                 & NICP$^*$                      & 0.34                       & 0.22                      & 3.43                     & 0.34                       & 0.21                      & 3.16                     & 0.35                       & 0.19                      & 2.76                     & 0.37                       & 0.19                      & 3.40                     & 0.37                       & 0.19                      & 3.40                     & 0.36                       & 0.22                      & 3.73                     \\
                                                 & DynaWeightPnP              & 0.46                       & 0.36                      & 5.00                     & 0.59                       & 0.38                      & 5.00                     & 0.59                       & 0.38                      & 5.00                     & 0.54                       & 0.36                      & 5.00                     & 0.52                       & 0.36                      & 5.00                     & 0.52                       & 0.36                      & 5.00                     \\
                                                 & DynaWeightPnP$^*$             & 0.35                       & 0.23                      & 3.85                     & 0.31                       & 0.18                      & 2.76                     & 0.34                       & 0.19                      & 2.76                     & 0.37                       & 0.19                      & 3.31                     & 0.36                       & 0.20                      & 3.45                     & 0.36                       & 0.22                      & 3.86                     \\ \midrule
\multirow{4}{*}{\makecell[l]{$8mm$,\\$3^\circ$}} & NICP                       & 0.69                       & 0.81                      & 8.00                     & 0.76                       & 0.62                      & 8.00                     & 1.06                       & 0.60                      & 8.00                     & 0.80                       & 0.60                      & 8.00                     & 0.76                       & 0.58                      & 8.00                     & 0.73                       & 0.58                      & 8.00                     \\
                                                 & NICP$^*$                      & 0.38                       & 0.30                      & 4.25                     & 0.36                       & 0.18                      & 3.09                     & 0.39                       & 0.18                      & 3.02                     & 0.41                       & 0.18                      & 3.72                     & 0.41                       & 0.22                      & 4.43                     & 0.40                       & 0.24                      & 4.43                     \\
                                                 & DynaWeightPnP              & 0.57                       & 0.54                      & 8.00                     & 0.84                       & 0.58                      & 8.00                     & 0.96                       & 0.58                      & 8.00                     & 0.75                       & 0.57                      & 8.00                     & 0.71                       & 0.57                      & 8.00                     & 0.69                       & 0.57                      & 8.00                     \\
                                                 & DynaWeightPnP$^*$             & 0.38                       & 0.27                      & 4.25                     & 0.37                       & 0.18                      & 3.02                     & 0.38                       & 0.18                      & 3.02                     & 0.43                       & 0.19                      & 3.97                     & 0.41                       & 0.23                      & 4.44                     & 0.40                       & 0.24                      & 4.43                     \\ \midrule
\multirow{4}{*}{\makecell[l]{$10mm$,\\$5^\circ$}} & NICP                       & 1.65                       & 1.56                      & 10.00                    & 1.39                       & 0.91                      & 10.00                    & 1.29                       & 0.77                      & 10.00                    & 1.09                       & 0.77                      & 10.00                    & 1.04                       & 0.75                      & 10.00                    & 1.03                       & 0.75                      & 10.00                    \\
                                                 & NICP$^*$                      & 0.43                       & 0.39                      & 5.66                     & 0.41                       & 0.29                      & 4.55                     & 0.41                       & 0.26                      & 4.08                     & 0.43                       & 0.26                      & 4.49                     & 0.43                       & 0.27                      & 4.67                     & 0.41                       & 0.28                      & 5.02                     \\
                                                 & DynaWeightPnP              & 0.82                       & 0.71                      & 10.00                    & 1.22                       & 0.75                      & 10.00                    & 1.12                       & 0.75                      & 10.00                    & 1.03                       & 0.74                      & 10.00                    & 1.01                       & 0.74                      & 10.00                    & 1.00                       & 0.74                      & 10.00                    \\
                                                 & DynaWeightPnP$^*$             & 0.39                       & 0.35                      & 4.65                     & 0.41                       & 0.27                      & 4.17                     & 0.40                       & 0.25                      & 4.09                     & 0.44                       & 0.27                      & 4.67                     & 0.41                       & 0.28                      & 4.79                     & 0.41                       & 0.28                      & 4.86                     \\ \bottomrule
\end{tabular}
}
\label{Table_IRLS_ADMM_nonoise}
\end{table*}

\begin{table*}[]
\caption{The table presents the results of NICP, NICP$^*$, DynaWeightPnP and DynaWeightPnP$^*$. NICP$^*$ denotes NICP with alternative searching. Then, their results are adopted to initialize the Algorithm 1 for post refinement. The simulation 2D projections are generated by enforcing 10 disturbances on the original pose. \textbf{Different from Table \ref{Table_IRLS_ADMM_nonoise}, 5 leafs of the simulated 2D vessel projection are randomly pruned.} Then, median TRE (``Median'') and pose difference  (``Dist.'' for translational difference and ``Angle'' for angular difference) are calculated for the 10 trials. All results are presented in average.}
\setlength{\tabcolsep}{1pt}
\resizebox{\columnwidth}{!}{
\begin{tabular}{llcccccccccccccccccc}
\toprule
\multirow{2}{*}{}                                 & \multirow{2}{*}{Algorithm} & \multicolumn{3}{c}{P\_1} & \multicolumn{3}{c}{P\_2} & \multicolumn{3}{c}{P\_3} & \multicolumn{3}{c}{P\_4} & \multicolumn{3}{c}{P\_5} & \multicolumn{3}{c}{P\_6} \\ \cline{3-20} 
                                                  &                            & Median   & Angle  & Dist   & Median   & Angle   & Dist   & Median   & Angle   & Dist   & Median   & Angle   & Dist   & Median   & Angle  & Dist   & Median   & Angle  & Dist   \\ \midrule
\multirow{4}{*}{\makecell[l]{$5mm$,\\$2^\circ$}}  & NICP                       & 0.60     & 0.59   & 5.00   & 0.59     & 0.38    & 5.00   & 0.67     & 0.38    & 5.00   & 0.58     & 0.38    & 5.00   & 0.52     & 0.37   & 5.00   & 0.52     & 0.37   & 5.00   \\
                                                  & NICP$^*$                      & 0.45     & 0.39   & 5.79   & 0.44     & 0.31    & 4.44   & 0.45     & 0.29    & 4.03   & 0.45     & 0.29    & 4.53   & 0.44     & 0.28   & 4.55   & 0.41     & 0.27   & 4.72   \\
                                                  & DynaWeightPnP              & 0.45     & 0.35   & 5.00   & 0.59     & 0.35    & 5.00   & 0.59     & 0.36    & 5.00   & 0.52     & 0.36    & 5.00   & 0.51     & 0.36   & 5.00   & 0.50     & 0.36   & 5.00   \\
                                                  & DynaWeightPnP$^*$             & 0.44     & 0.33   & 5.00   & 0.44     & 0.27    & 3.76   & 0.44     & 0.30    & 4.37   & 0.44     & 0.28    & 4.61   & 0.41     & 0.28   & 4.69   & 0.40     & 0.28   & 5.00   \\ \midrule
\multirow{4}{*}{\makecell[l]{$8mm$,\\$3^\circ$}}  & NICP                       & 0.72     & 0.71   & 8.00   & 0.75     & 0.59    & 8.00   & 0.88     & 0.59    & 8.00   & 0.78     & 0.58    & 8.00   & 0.75     & 0.58   & 8.00   & 0.70     & 0.58   & 8.00   \\
                                                  & NICP$^*$                      & 0.50     & 0.44   & 5.42   & 0.53     & 0.38    & 5.46   & 0.57     & 0.36    & 5.42   & 0.53     & 0.32    & 5.35   & 0.51     & 0.30   & 5.35   & 0.49     & 0.32   & 5.44   \\
                                                  & DynaWeightPnP              & 0.60     & 0.54   & 8.00   & 0.82     & 0.58    & 8.00   & 0.82     & 0.58    & 8.00   & 0.69     & 0.57    & 8.00   & 0.69     & 0.57   & 8.00   & 0.66     & 0.57   & 8.00   \\
                                                  & DynaWeightPnP$^*$        & 0.47     & 0.38   & 5.62   & 0.54     & 0.37    & 5.44   & 0.57     & 0.35    & 5.40   & 0.53     & 0.31    & 5.40   & 0.49     & 0.30   & 5.38   & 0.48     & 0.32   & 5.79   \\ \midrule
\multirow{4}{*}{\makecell[l]{$10mm$,\\$5^\circ$}} & NICP                       & 1.65     & 1.56   & 10.00  & 1.39     & 0.91    & 10.00  & 1.29     & 0.77    & 10.00  & 1.09     & 0.77    & 10.00  & 1.04     & 0.75   & 10.00  & 1.03     & 0.75   & 10.00  \\
                                                  & NICP$^*$                      & 0.43     & 0.39   & 5.66   & 0.41     & 0.29    & 4.55   & 0.41     & 0.26    & 4.08   & 0.43     & 0.26    & 4.49   & 0.43     & 0.27   & 4.67   & 0.41     & 0.28   & 5.02   \\
                                                  & DynaWeightPnP              & 0.82     & 0.71   & 10.00  & 1.22     & 0.75    & 10.00  & 1.12     & 0.75    & 10.00  & 1.03     & 0.74    & 10.00  & 1.01     & 0.74   & 10.00  & 1.01     & 0.74   & 10.00  \\
                                                  & DynaWeightPnP$^*$             & 0.39     & 0.35   & 4.65   & 0.41     & 0.27    & 4.17   & 0.40     & 0.25    & 4.09   & 0.44     & 0.27    & 4.67   & 0.41     & 0.28   & 4.79   & 0.41     & 0.28   & 4.86   \\ \bottomrule
\end{tabular}
}
\label{Table_IRLS_ADMM_noise}
\end{table*}

	\section{Conclusion}
	\label{section_conclusion}
	
	This paper explores the correspondence-free PnP, that is, estimating the optimal pose to align 3D and 2D shapes in real-time without correspondences. We reveal two main problems in correspondence-free PnP, that is \textbf{partially overlap} and \textbf{numerical observability} issue between rotation and translation. Different from previous methods with Euclidean distance and Huber loss, this research introduces the RKHS to tackle the partial overlap issue. An IRLS method is then employed to solve the RKHS-based formulation for real-time implementation efficiently. Additionally, our research first discover and analyzes a numerical ambiguity between rotation and translation in correspondence-free PnP. The observability issue causes a massive amount of local minimum in contrast to conventional PnP, which maintains good correspondences. To address this issue, we introduce the RKHS formulation to improve DynaWeightPnP, which features a dynamic weighting sub-problem and an alternative searching algorithm designed as a post-refinement to improve pose estimation and alignment accuracy.\par

    Experiments were conducted on a typical case, specifically a 3D-2D vascular centerline registration task within EIGIs. The results from the in-vivo and ex-vivo experiments demonstrate that the proposed DynaWeightPnP achieves registration processing rates of 60 Hz (without dynamic weighted searching) and 31 Hz (with dynamic weighted searching) on modern single-core CPU, with accuracy comparable to existing methods. These findings underscore the potential of DynaWeightPnP for future robotic navigation tasks such as EIGIs.

    {\small
		\bibliographystyle{elsarticle-num-names}
		\bibliography{strings-abrv,ieee-abrv,reference}

@STRING{IEEE_J_CASVT      = "{IEEE} Trans. Circuits Syst. Video Technol."}

@STRING{IEEE_J_VCG        = "{IEEE} Trans. Vis. Comput. Graphics"}

@STRING{IEEE_J_PAMI       = "{IEEE} Trans. Pattern Anal. Mach. Intell."}

@STRING{IEEE_J_RO         = "{IEEE} Trans. Robot."}

@STRING{IEEE_J_SMC        = "{IEEE} Trans. Syst., Man, Cybern."}

@STRING{IEEE_J_GRS        = "{IEEE} Trans. Geosci. Remote Sens."}

@STRING{IEEE_J_MI         = "{IEEE} Trans. Med. Imag."}

@STRING{IEEE_J_MRB         = "{IEEE} Trans. on Med. Rob. and Bio."}

@STRING{IEEE_J_RAL         = "{IEEE} Robot. and Autom. Lett."}

@book{boumal2023introduction,
  title={An introduction to optimization on smooth manifolds},
  author={Boumal, Nicolas},
  year={2023},
  publisher={Cambridge University Press}
}

@article{rivest2012nonrigid,
  title={Nonrigid {2D/3D} registration of coronary artery models with live fluoroscopy for guidance of cardiac interventions},
  author={Rivest-Henault, David and Sundar, Hari and Cheriet, Mohamed},
  journal=IEEE_J_MI,
  volume={31},
  number={8},
  pages={1557--1572},
  year={2012},
  publisher={IEEE}
}

@inproceedings{jomier20063d,
  title={{3D/2D} model-to-image registration applied to {TIPS} surgery},
  author={Jomier, Julien and Bullitt, Elizabeth and Horn, Mark Van and Pathak, Chetna and Aylward, Stephen R},
  booktitle=C-MICCAI,
  pages={662--669},
  year={2006},
  organization={Springer}
}

@article{aylward2003registration,
  title={Registration and analysis of vascular images},
  author={Aylward, Stephen R and Jomier, Julien and Weeks, Sue and Bullitt, Elizabeth},
  journal=J-IJCV,
  volume={55},
  number={2},
  pages={123--138},
  year={2003},
  publisher={Springer}
}

@article{mitrovic20183d,
  title={{3D--2D} registration in endovascular image-guided surgery: Evaluation of state-of-the-art methods on cerebral angiograms},
  author={Mitrovi{\'c}, Uro{\v{s}} and Likar, Bo{\v{s}}tjan and Pernu{\v{s}}, Franjo and {\v{S}}piclin, {\v{Z}}iga},
  journal=J-CARS,
  volume={13},
  number={2},
  pages={193--202},
  year={2018},
  publisher={Springer}
}

@book{hartley2003multiple,
  title={Multiple view geometry in computer vision},
  author={Hartley, Richard and Zisserman, Andrew},
  year={2003},
  publisher={Cambridge university press}
}

@inproceedings{triggs2000bundle,
  title={Bundle adjustment—a modern synthesis},
  author={Triggs, Bill and McLauchlan, Philip F and Hartley, Richard I and Fitzgibbon, Andrew W},
  booktitle={Vision algorithms: theory and practice},
  pages={298--372},
  year={2000},
  organization={Springer}
}

@article{zheng2018pairwise,
  title={Pairwise domain adaptation module for {CNN-based 2-D/3-D} registration},
  author={Zheng, Jiannan and Miao, Shun and Jane Wang, Z and Liao, Rui},
  journal={Journal of Medical Imaging},
  volume={5},
  number={2},
  pages={021204--021204},
  year={2018},
  publisher={Society of Photo-Optical Instrumentation Engineers}
}

@inproceedings{liao2017artificial,
  title={An artificial agent for robust image registration},
  author={Liao, Rui and Miao, Shun and de Tournemire, Pierre and Grbic, Sasa and Kamen, Ali and Mansi, Tommaso and Comaniciu, Dorin},
  booktitle=C-AAAI,
  volume={31},
  number={1},
  year={2017}
}

@inproceedings{miao2018dilated,
  title={Dilated {FCN} for multi-agent {2D/3D} medical image registration},
  author={Miao, Shun and Piat, Sebastien and Fischer, Peter and Tuysuzoglu, Ahmet and Mewes, Philip and Mansi, Tommaso and Liao, Rui},
  booktitle=C-AAAI,
  volume={32},
  number={1},
  year={2018}
}

@article{guan2020transfer,
  title={Transfer learning for nonrigid {2D/3D} cardiovascular images registration},
  author={Guan, Shaoya and Wang, Tianmiao and Sun, Kai and Meng, Cai},
  journal={IEEE Journal of Biomedical and Health Informatics},
  volume={25},
  number={9},
  pages={3300--3309},
  year={2020},
  publisher={IEEE}
}

@article{guan2019deformable,
  title={Deformable cardiovascular image registration via multi-channel convolutional neural network},
  author={Guan, Shaoya and Meng, Cai and Xie, Yi and Wang, Qi and Sun, Kai and Wang, Tianmiao},
  journal={IEEE Access},
  volume={7},
  pages={17524--17534},
  year={2019},
  publisher={IEEE}
}

@article{groher2009deformable,
  title={Deformable {2D-3D} registration of vascular structures in a one view scenario},
  author={Groher, Martin and Zikic, Darko and Navab, Nassir},
  journal=IEEE_J_MI,
  volume={28},
  number={6},
  pages={847--860},
  year={2009},
  publisher={IEEE}
}

@article{mitrovic20133d,
  title={{3D-2D} Registration of Cerebral Angiograms: A Method and Evaluation on Clinical Images},
  author={Mitrovic, Uros and Spiclin, Ziga and Likar, Bostjan and Pernus, Franjo},
  journal=IEEE_J_MI,
  volume={8},
  number={32},
  pages={1550--1563},
  year={2013}
}

@inproceedings{zheng2017learning,
  title={Learning {CNNS} with pairwise domain adaption for real-time 6{DoF} ultrasound transducer detection and tracking from {X}-ray images},
  author={Zheng, Jiannan and Miao, Shun and Liao, Rui},
  booktitle=C-MICCAI,
  pages={646--654},
  year={2017},
  organization={Springer}
}

@article{miao2019agent,
  title={Agent-based methods for medical image registration},
  author={Miao, Shun and Liao, Rui},
  journal={Deep learning and convolutional neural networks for medical imaging and clinical informatics},
  pages={323--345},
  year={2019},
  publisher={Springer}
}

@article{otsu1979threshold,
  title={A threshold selection method from gray-level histograms},
  author={Otsu, Nobuyuki},
  journal=IEEE_J_SMC,
  volume={9},
  number={1},
  pages={62--66},
  year={1979},
  publisher={IEEE}
}

@article{li2022dual,
  title={Dual encoder-based dynamic-channel graph convolutional network with edge enhancement for retinal vessel segmentation},
  author={Li, Yang and Zhang, Yue and Cui, Weigang and Lei, Baiying and Kuang, Xihe and Zhang, Teng},
  journal=IEEE_J_MI,
  volume={41},
  number={8},
  pages={1975--1989},
  year={2022},
  publisher={IEEE}
}

@article{moccia2018blood,
  title={Blood vessel segmentation algorithms—review of methods, datasets and evaluation metrics},
  author={Moccia, Sara and De Momi, Elena and El Hadji, Sara and Mattos, Leonardo S},
  journal=J-CMPB,
  volume={158},
  pages={71--91},
  year={2018},
  publisher={Elsevier}
}

@article{fischler1981random,
  title={Random sample consensus: a paradigm for model fitting with applications to image analysis and automated cartography},
  author={Fischler, Martin A and Bolles, Robert C},
  journal={Communications of the ACM},
  volume={24},
  number={6},
  pages={381--395},
  year={1981},
  publisher={ACM New York, NY, USA}
}

@article{clark2021nonparametric,
  title={Nonparametric continuous sensor registration},
  author={Clark, William and Ghaffari, Maani and Bloch, Anthony},
  journal=J-JMLR,
  volume={22},
  number={1},
  pages={12412--12461},
  year={2021},
  publisher={JMLRORG}
}

@inproceedings{biber2003normal,
  title={The normal distributions transform: A new approach to laser scan matching},
  author={Biber, Peter and Stra{\ss}er, Wolfgang},
  booktitle=C-IROS,
  volume={3},
  pages={2743--2748},
  year={2003},
  organization={IEEE}
}

@article{magnusson2007scan,
  title={Scan registration for autonomous mining vehicles using {3D-NDT}},
  author={Magnusson, Martin and Lilienthal, Achim and Duckett, Tom},
  journal=J-JFR,
  volume={24},
  number={10},
  pages={803--827},
  year={2007},
  publisher={Wiley Online Library}
}

@article{jian2010robust,
  title={Robust point set registration using gaussian mixture models},
  author={Jian, Bing and Vemuri, Baba C},
  journal=IEEE_J_PAMI,
  volume={33},
  number={8},
  pages={1633--1645},
  year={2010},
  publisher={IEEE}
}

@article{markelj2008robust,
  title={Robust gradient-based {3-D/2-D} registration of {CT} and {MR} to {X}-ray images},
  author={Markelj, Primoz and Tomazevic, Dejan and Pernus, Franjo and Likar, Bostjan},
  journal=IEEE_J_MI,
  volume={27},
  number={12},
  pages={1704--1714},
  year={2008},
  publisher={IEEE}
}

@article{meng2022weakly,
  title={A weakly supervised framework for {2D/3D} vascular registration oriented to incomplete {2D} blood vessels},
  author={Meng, Cai and Li, Yanggang and Xu, Yizhou and Li, Ning and Xia, Kun},
  journal=IEEE_J_MRB,
  volume={4},
  number={2},
  pages={381--390},
  year={2022},
  publisher={IEEE}
}

@inproceedings{quigley2009ros,
  title={{ROS}: an open-source Robot Operating System},
  author={Quigley, Morgan and Conley, Ken and Gerkey, Brian and Faust, Josh and Foote, Tully and Leibs, Jeremy and Wheeler, Rob and Ng, Andrew Y and others},
  booktitle={ICRA workshop on open source software},
  volume={3},
  number={3.2},
  pages={5},
  year={2009},
  organization={Kobe, Japan}
}

@article{maronna1976robust,
  title={Robust {M}-estimators of multivariate location and scatter},
  author={Maronna, Ricardo Antonio},
  journal={The annals of statistics},
  pages={51--67},
  year={1976},
  publisher={JSTOR}
}

@article{haralick1994review,
  title={Review and analysis of solutions of the three point perspective pose estimation problem},
  author={Haralick, Bert M and Lee, Chung-Nan and Ottenberg, Karsten and N{\"o}lle, Michael},
  journal=J-IJCV,
  volume={13},
  pages={331--356},
  year={1994},
  publisher={Springer}
}

@article{Lepetit_160138,
      title = {{EPnP}: {An} Accurate {O(n)} Solution to the {PnP} Problem},
      author = {Lepetit, Vincent and Moreno-Noguer, Francesc and Fua,  Pascal},
      publisher = {Springer Verlag},
      journal = J-IJCV,
      volume = {81},
      pages = {155-166},
      year = {2009},
      url = {http://infoscience.epfl.ch/record/160138},
      doi = {https://doi.org/10.1007/s11263-008-0152-6},
}

@article{steininger2012auto,
  title={Auto-masked {2D/3D} image registration and its validation with clinical cone-beam computed tomography},
  author={Steininger, P and Neuner, M and Weichenberger, H and Sharp, GC and Winey, B and Kametriser, G and Sedlmayer, F and Deutschmann, H},
  journal={Physics in Medicine And Biology},
  volume={57},
  number={13},
  pages={4277},
  year={2012},
  publisher={IOP Publishing}
}

@article{zitova2003image,
  title={Image registration methods: {A} survey},
  author={Zitova, Barbara and Flusser, Jan},
  journal=J-IVC,
  volume={21},
  number={11},
  pages={977--1000},
  year={2003},
  publisher={Elsevier}
}

@INPROCEEDINGS{song2023iterative,
  author={Song, Jingwei and Yang, Keke and Zhang, Zheng and Li, Meng and Cao, Tuoyu and Ghaffari, Maani},
  booktitle=C-ICRA, 
  title={Iterative PnP and its application in 3D-2D vascular image registration for robot navigation}, 
  year={2024},
  volume={},
  number={},
  pages={17560-17566},
  organization={IEEE}}

@inproceedings{besl1992method,
  title={Method for registration of {3-D} shapes},
  author={Besl, Paul J and McKay, Neil D},
  booktitle={Sensor fusion IV: control paradigms and data structures},
  volume={1611},
  pages={586--606},
  year={1992},
  organization={Spie}
}

@article{quan1999linear,
  title={Linear n-point camera pose determination},
  author={Quan, Long and Lan, Zhongdan},
  journal=IEEE_J_PAMI,
  volume={21},
  number={8},
  pages={774--780},
  year={1999},
  publisher={IEEE}
}

@article{lepetit2009ep,
  title={{EPnP}: {An} accurate {O(n)} solution to the {PnP} problem},
  author={Lepetit, Vincent and Moreno-Noguer, Francesc and Fua, Pascal},
  journal=J-IJCV,
  volume={81},
  pages={155--166},
  year={2009},
  publisher={Springer}
}

@article{arnold2021fast,
  title={Fast and robust registration of partially overlapping point clouds},
  author={Arnold, Eduardo and Mozaffari, Sajjad and Dianati, Mehrdad},
  journal=IEEE_J_RAL,
  volume={7},
  number={2},
  pages={1502--1509},
  year={2021},
  publisher={IEEE}
}

@article{gao2003complete,
  title={Complete solution classification for the perspective-three-point problem},
  author={Gao, Xiao-Shan and Hou, Xiao-Rong and Tang, Jianliang and Cheng, Hang-Fei},
  journal=IEEE_J_PAMI,
  volume={25},
  number={8},
  pages={930--943},
  year={2003},
  publisher={IEEE}
}

@inproceedings{kneip2011novel,
  title={A novel parametrization of the perspective-three-point problem for a direct computation of absolute camera position and orientation},
  author={Kneip, Laurent and Scaramuzza, Davide and Siegwart, Roland},
  booktitle=C-CVPR,
  pages={2969--2976},
  year={2011},
  organization={IEEE}
}

@article{hu2002note,
  title={A note on the number of solutions of the noncoplanar {P4P} problem},
  author={Hu, ZY and Wu, FC},
  journal=IEEE_J_PAMI,
  volume={24},
  number={4},
  pages={550--555},
  year={2002},
  publisher={IEEE}
}

@article{li2012robust,
  title={A robust {O(n)} solution to the perspective-n-point problem},
  author={Li, Shiqi and Xu, Chi and Xie, Ming},
  journal=IEEE_J_PAMI,
  volume={34},
  number={7},
  pages={1444--1450},
  year={2012},
  publisher={IEEE}
}

@book{berlinet2011reproducing,
  title={Reproducing kernel Hilbert spaces in probability and statistics},
  author={Berlinet, Alain and Thomas-Agnan, Christine},
  year={2011},
  publisher={Springer Science \& Business Media}
}

@book{mohri2018foundations,
  title={Foundations of machine learning},
  author={Mohri, Mehryar and Rostamizadeh, Afshin and Talwalkar, Ameet},
  year={2018},
  publisher={MIT press}
}

@article{ray2024rkhsba,
  title={{RKHS-BA}: A Semantic Correspondence-Free Multi-View Registration Framework with Global Tracking},
  author={Zhang, Ray and Song, Jingwei and Gao, Xiang and Wu, Junzhe and Liu, Tianyi and Zhang, Jinyuan and Eustice, Ryan and Ghaffari, Maani},
  journal={arXiv preprint arXiv:2403.01254},
  year={2024}
}

@book{steinwart2008support,
  title={Support vector machines},
  author={Steinwart, Ingo and Christmann, Andreas},
  year={2008},
  publisher={Springer Science \& Business Media}
}

@book{bar2004estimation,
  title={Estimation with applications to tracking and navigation: theory algorithms and software},
  author={Bar-Shalom, Yaakov and Li, X Rong and Kirubarajan, Thiagalingam},
  year={2004},
  publisher={John Wiley \& Sons}
}

@article{jiang2021review,
  title={A review of multimodal image matching: Methods and applications},
  author={Jiang, Xingyu and Ma, Jiayi and Xiao, Guobao and Shao, Zhenfeng and Guo, Xiaojie},
  journal={Information Fusion},
  volume={73},
  pages={22--71},
  year={2021},
  publisher={Elsevier}
}

@article{pore2023autonomous,
  title={Autonomous navigation for robot-assisted intraluminal and endovascular procedures: A systematic review},
  author={Pore, Ameya and Li, Zhen and Dall'Alba, Diego and Hernansanz, Albert and De Momi, Elena and Menciassi, Arianna and Gelpi, Alicia Casals and Dankelman, Jenny and Fiorini, Paolo and Vander Poorten, Emmanuel},
  journal=IEEE_J_RO,
  volume={39},
  number={4},
  pages={2529--2548},
  year={2023},
  publisher={IEEE}
}

@article{labrunie2022automatic,
  title={Automatic preoperative {3D} model registration in laparoscopic liver resection},
  author={Labrunie, Mathieu and Ribeiro, Mathieu and Mourthadhoi, Farouk and Tilmant, Christophe and Le Roy, Bertrand and Buc, Emmanuel and Bartoli, Adrien},
  journal=J-CARS,
  volume={17},
  number={8},
  pages={1429--1436},
  year={2022},
  publisher={Springer}
}

@article{wang2020multientity,
  title={Multientity registration of point clouds for dynamic objects on complex floating platform using object silhouettes},
  author={Wang, Feng and Hu, Han and Ge, Xuming and Xu, Bo and Zhong, Ruofei and Ding, Yulin and Xie, Xiao and Zhu, Qing},
  journal=IEEE_J_GRS,
  number={1},
  pages={769--783},
  year={2020},
  publisher={IEEE}
}

@article{perez2024alignment,
  title={Alignment and improvement of shape-from-silhouette reconstructed 3D objects},
  author={Perez, Alberto J and Perez-Soler, Javier and Perez-Cortes, Juan-Carlos and Guardiola, Jose-Luis},
  journal={IEEE Access},
  year={2024},
  publisher={IEEE}
}

@inproceedings{tian20203d,
  title={3D scene geometry-aware constraint for camera localization with deep learning},
  author={Tian, Mi and Nie, Qiong and Shen, Hao},
  booktitle=C-ICRA,
  pages={4211--4217},
  year={2020},
  organization={IEEE}
}

@article{song2022fusing,
  title={Fusing convolutional neural network and geometric constraint for image-based indoor localization},
  author={Song, Jingwei and Patel, Mitesh and Ghaffari, Maani},
  journal=IEEE_J_RAL,
  volume={7},
  number={2},
  pages={1674--1681},
  year={2022},
  publisher={IEEE}
}

@article{muja2009flann,
  title={Flann-fast library for approximate nearest neighbors user manual},
  author={Muja, Marius and Lowe, David},
  journal={Computer Science Department, University of British Columbia, Vancouver, BC, Canada},
  volume={5},
  number={6},
  year={2009}
}

@inproceedings{harris1990rapid,
  title={{RAPID-a} video rate object tracker.},
  author={Harris, Chris and Stennett, Carl},
  booktitle=C-BMVC,
  pages={1--6},
  year={1990}
}

@inproceedings{choi20123d,
  title={{3D} textureless object detection and tracking: An edge-based approach},
  author={Choi, Changhyun and Christensen, Henrik I},
  booktitle=C-IROS,
  pages={3877--3884},
  year={2012},
  organization={IEEE}
}

@article{seo2013optimal,
  title={Optimal local searching for fast and robust textureless {3D} object tracking in highly cluttered backgrounds},
  author={Seo, Byung-Kuk and Park, Hanhoon and Park, Jong-Il and Hinterstoisser, Stefan and Ilic, Slobodan},
  journal=IEEE_J_VCG,
  volume={20},
  number={1},
  pages={99--110},
  year={2013},
  publisher={IEEE}
}

@inproceedings{tian2022large,
  title={Large-displacement {3D} object tracking with hybrid non-local optimization},
  author={Tian, Xuhui and Lin, Xinran and Zhong, Fan and Qin, Xueying},
  booktitle=C-ECCV,
  pages={627--643},
  year={2022},
  organization={Springer}
}

@article{dong2020accurate,
  title={Accurate {6DOF} pose tracking for texture-less objects},
  author={Dong, Yanchao and Ji, Lingling and Wang, Senbo and Gong, Pei and Yue, Jiguang and Shen, Runjie and Chen, Ce and Zhang, Yaping},
  journal=IEEE_J_CASVT,
  volume={31},
  number={5},
  pages={1834--1848},
  year={2020},
  publisher={IEEE}
}

@article{drummond2002real,
  title={Real-time visual tracking of complex structures},
  author={Drummond, Tom and Cipolla, Roberto},
  journal=IEEE_J_PAMI,
  volume={24},
  number={7},
  pages={932--946},
  year={2002},
  publisher={IEEE}
}

@article{huang2021pixel,
  title={Pixel-wise weighted region-based 3D object tracking using contour constraints},
  author={Huang, Hong and Zhong, Fan and Qin, Xueying},
  journal=IEEE_J_VCG,
  volume={28},
  number={12},
  pages={4319--4331},
  year={2021},
  publisher={IEEE}
}

@STRING{C-AAAI = {Proc. {AAAI} Nat. Conf. Artif. Intell.}}

@STRING{C-BMVC = {Proc. British Mach. Vis. Conf.}}

@STRING{C-CVPR = {Proc. {IEEE} Conf. Comput. Vis. Pattern Recog.}}

@STRING{C-ECCV = {Proc. European Conf. Comput. Vis.}}

@STRING{C-ICRA = {Proc. {IEEE} Int. Conf. Robot. and Automation}}

@STRING{C-IROS = {Proc. {IEEE}/{RSJ} Int. Conf. Intell. Robots and Syst.}}

@STRING{C-MICCAI = { Int. Conf. on Med. Image Comput. and Comput. Assist. Interv.}}

@STRING{J-IJCV = {Int. J. Comput. Vis.}}

@STRING{J-IVC = {Image and Vis. Comput.}}

@STRING{J-JFR = {J. Field Robot.}}

@STRING{J-JMLR = {J. Mach. Learning Res.}}

@STRING{J-CMPB = {Comp. meth. and prog. in biomed.}}

@STRING{J-CARS = {Int. Jou. of Comp. Assis. Rad. and Surg.}}
	}
    



\end{document}